%% file: MIRAGE.tex
\pdfoutput=1

\documentclass[11pt]{article}

\usepackage{acl}

\usepackage{times}
\usepackage{amsmath}
\usepackage{latexsym}
\usepackage{adjustbox}
\usepackage{float}
\usepackage{makecell}
\usepackage{multirow}
\usepackage[cjk]{kotex}
\usepackage{array}
\usepackage{booktabs}
\usepackage{graphicx}
\usepackage{array}
\usepackage{colortbl}
\usepackage{tcolorbox}
\usepackage{booktabs}
\usepackage{tabularx}
\usepackage{hyperref}

\usepackage[T1]{fontenc}

\usepackage[utf8]{inputenc}

\usepackage{microtype}

\usepackage{inconsolata}

\usepackage{graphicx}

\title{MIRAGE: A Metric-Intensive Benchmark
\\ for Retrieval-Augmented Generation Evaluation}

\author{Chanhee Park, Hyeonseok Moon, Chanjun Park$^{\dagger}$, Heuiseok Lim$^{\dagger}$ \\
  Korea University, Republic of Korea \\
  \texttt{\{pch7678, glee889, bcj1210, limhseok\}@korea.ac.kr} \\}

\newcommand\blfootnote[1]{%
  \begingroup
  \renewcommand\thefootnote{}\footnote{#1}%
  \addtocounter{footnote}{-1}%
  \endgroup
}

\begin{document}
\maketitle
\begin{abstract}
Retrieval-Augmented Generation (RAG) has gained prominence as an effective method for enhancing the generative capabilities of Large Language Models (LLMs) through the incorporation of external knowledge. However, the evaluation of RAG systems remains a challenge, due to the intricate interplay between retrieval and generation components. This limitation has resulted in a scarcity of benchmarks that facilitate a detailed, component-specific assessment. In this work, we present MIRAGE, a Question Answering dataset specifically designed for RAG evaluation. MIRAGE consists of 7,560 curated instances mapped to a retrieval pool of 37,800 entries, enabling an efficient and precise evaluation of both retrieval and generation tasks. We also introduce novel evaluation metrics aimed at measuring RAG adaptability, encompassing dimensions such as noise vulnerability, context acceptability, context insensitivity, and context misinterpretation. Through comprehensive experiments across various retriever-LLM configurations, we provide new insights into the optimal alignment of model pairs and the nuanced dynamics within RAG systems. The dataset and evaluation code are publicly available, allowing for seamless integration and customization in diverse research settings\footnote{The MIRAGE code and data are available at \href{https://github.com/nlpai-lab/MIRAGE}{https://github.com/nlpai-lab/MIRAGE}.}.
\end{abstract}


\section{Introduction}
\blfootnote{$^\dagger$ Corresponding Author}
\input{Sections/1.Introduction}

\begin{figure*}[ht]
\centering
\includegraphics[width=\textwidth]{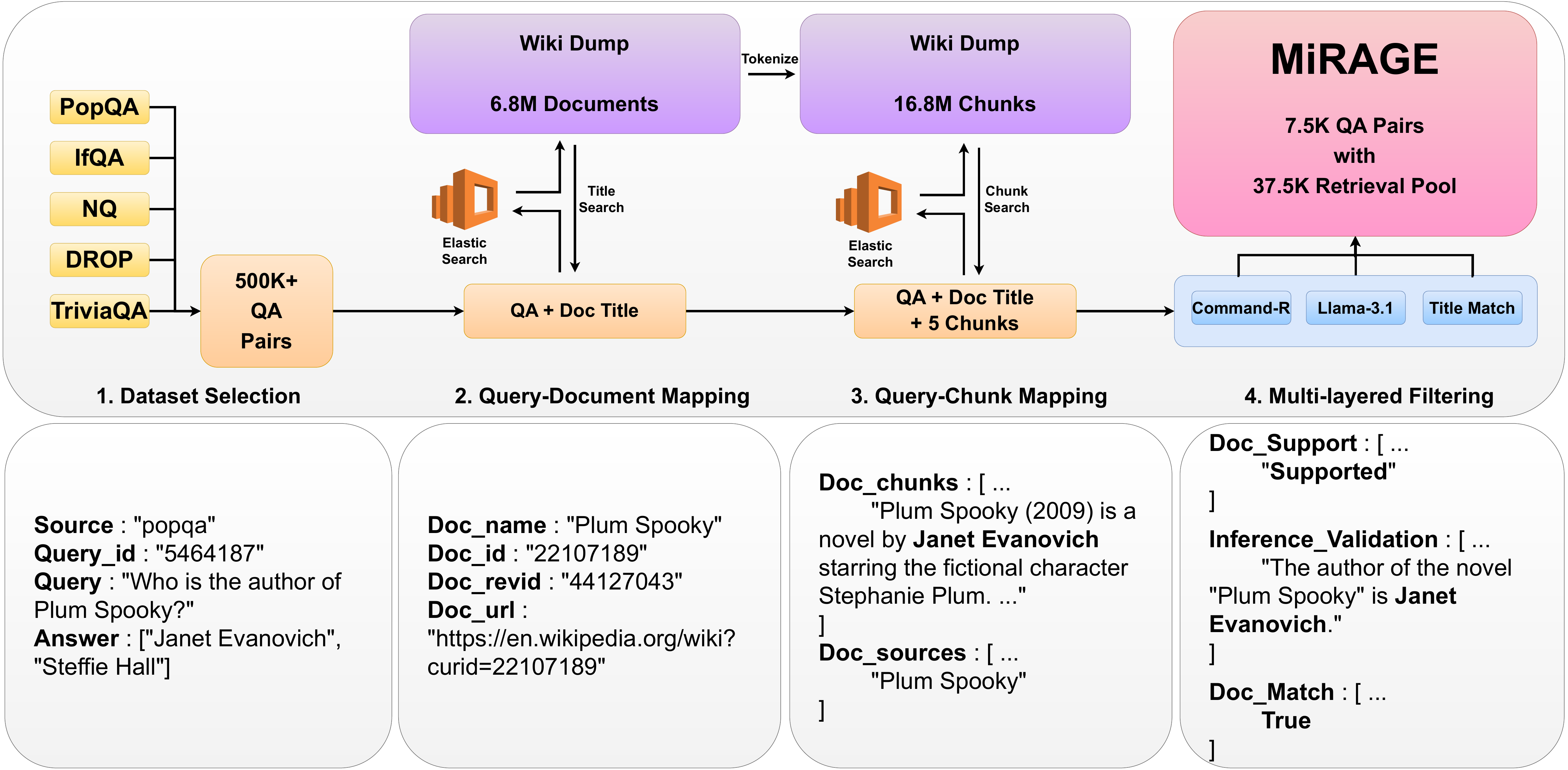}
\caption{Data Filtering Process for MIRAGE}
\label{fig:filtering}
\end{figure*}

\section{Related Work}

\input{Sections/2.Related_Works}

\section{MIRAGE}

\input{Sections/3.MiRAGE}

\section{Evaluation Framework}

\input{Sections/4.Evaluation_Measure}

\section{Experiments}

\input{Sections/5.Experiments}

\section{Conclusion}
In this paper, we introduce MIRAGE, a benchmark tailored to comprehensively evaluate the performance of RAG systems. Through extensive experiments, we demonstrate MIRAGE capability to provide detailed insights into the interaction between retrieval models and LLMs, revealing strengths and weaknesses in handling noisy contexts and incorporating external knowledge. MIRAGE addresses gaps left by existing benchmarks, offering a flexible framework for assessing RAG systems across various configurations, retrievers, and LLMs.

\section*{Limitations}
While MIRAGE contributes significant advancements to RAG system evaluation, several limitations warrant attention for future work:

\paragraph{Data Contamination Risk:} 
The potential for data contamination exists due to the use of publicly available datasets in constructing MIRAGE. Models evaluated on MIRAGE may have been exposed to parts of the dataset during pre-training or fine-tuning, leading to less accurate assessments. While we employed careful dataset partitioning and filtering to minimize this risk, complete elimination is challenging. Future iterations of MIRAGE should explore stricter partitioning strategies, such as temporal splits, to ensure that no overlap occurs between training and evaluation data.

\paragraph{Single-Hop Task Focus:} 
Although MIRAGE is intentionally designed to test multiple systems with minimum computational resources, its lightweight features are restricted to single-hop question answering, where answers are derived from a single oracle chunk. This simplifies the evaluation process but limits the complexity of tasks that require multi-hop reasoning. In real-world scenarios, models often need to integrate information from multiple sources. To better capture the complexity of real-world applications, future versions of MIRAGE should incorporate multi-hop tasks that require deeper reasoning across multiple document chunks.

\paragraph{Data Imbalance:} 
An inherent data imbalance exists across the QA pairs from different source datasets in MIRAGE. Certain datasets are more heavily represented than others, which could bias the evaluation results, particularly for retriever models that may learn to exploit frequent patterns. Addressing this imbalance in future versions of MIRAGE would allow for a more uniform evaluation of retrievers, ensuring that no specific dataset disproportionately influences the results.

\paragraph{Difficulty Level:} 
State-of-the-art models achieve high performance on MIRAGE, with some exceeding 90\% accuracy in oracle-based settings. While this indicates that MIRAGE effectively evaluates retrieval and generation, the benchmark may not be sufficiently challenging for oracle setups. However, in more complex, noisy settings, performance drops suggest that there is room for improvement, especially in handling ambiguous or noisy contexts. Future work could introduce more nuanced adversarial examples or task complexities to further increase the challenge level.

\paragraph{False Labels:} 
Despite rigorous filtering and human validation, a small proportion of false labels may exist in the MIRAGE dataset. In some instances, oracle chunks may have been incorrectly tagged or the answer labels may not perfectly align with the true answer. While these errors only affect a small portion of the dataset, they can still introduce noise into the evaluation. Future improvements to the labeling process should focus on reducing these errors to ensure a more robust dataset.

\section*{Ethics Statement}
The MIRAGE benchmark is built using publicly available datasets and resources, all of which comply with open-access policies. We ensured that no sensitive or private data was used during the construction of MIRAGE. Additionally, we emphasize the importance of data transparency and model accountability in our evaluation framework. While MIRAGE primarily focuses on technical evaluation, the ethical implications of model deployment in real-world applications should not be overlooked. We encourage users of MIRAGE to consider the societal impacts, potential biases, and fairness issues when deploying RAG systems evaluated using this benchmark. The dataset is intended for research purposes only, and care should be taken to ensure that the systems built upon it are deployed responsibly.






\section*{Acknowledgments}
This work was supported by ICT Creative Consilience Program through the Institute of Information \& Communications Technology Planning \& Evaluation(IITP) grant funded by the Korea government(MSIT)(IITP-2025-RS-2020-II201819). This work was supported by Institute for Information \& communications Technology Promotion(IITP) grant funded by the Korea government(MSIT) (RS-2024-00398115, Research on the reliability and coherence of outcomes produced by Generative AI). This work was supported by Institute of Information \& communications Technology Planning \& Evaluation(IITP) under the Leading Generative AI Human Resources Development(IITP-2024-R2408111) grant funded by the Korea government(MSIT).

\bibliography{custom}

\appendix

\section{Details of Model Prompts}
\label{sec:appendix_A}
\input{Tables/tb_prompt_base}

\input{Tables/tb_prompt_oracle}

\input{Tables/tb_prompt_mixed}

Table \ref{tab:base_prompt}, \ref{tab:oracle_prompt}, and \ref{tab:mixed_prompt} present the prompts used for the base, oracle, and mixed setups, respectively. We employed simple prompts to minimize the impact of instructions and to focus on evaluating the models' performance. These prompts are applied across all models utilized in the experiment. For the llama-3.1-8B-Instruct model during the validation process in \ref{sec:val}, each chunk is given one by one same as in the oracle setup.

\input{Tables/tb_prompt_command}

For the support label extraction process described in Section \ref{sec:sup}, we used the prompt shown in Table \ref{tab:sup_prompt}. Labels were extracted for all 37,800 chunks mapped to 7,560 queries to determine the relevance of each mapped chunk. To avoid interaction effects between chunks, each chunk was evaluated independently."

\section{Details of Data Distribution}
\label{sec:appendix_B}
\begin{figure}[ht]
\centering
\includegraphics[width=\columnwidth]{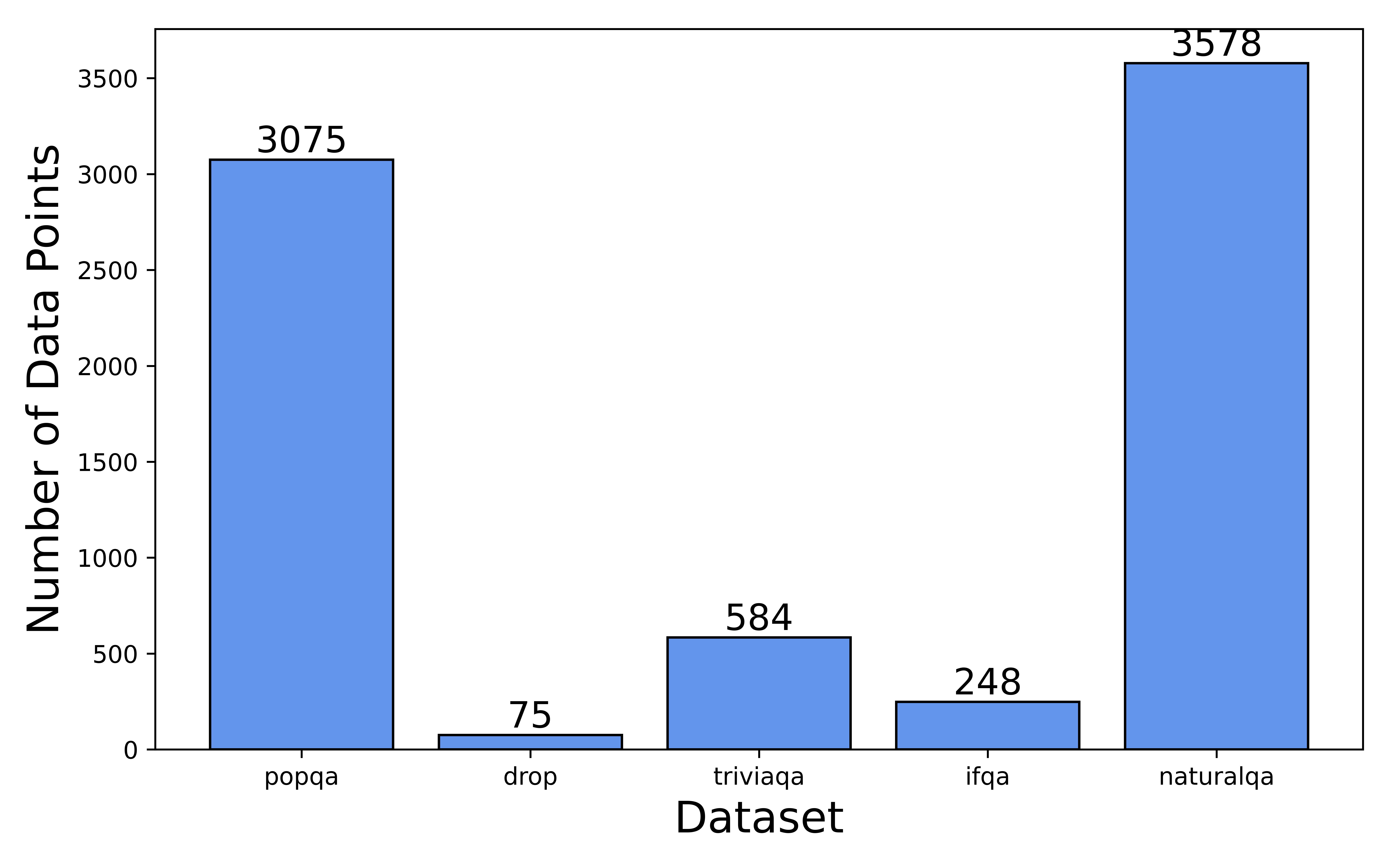}
\caption{Number of data points per dataset}
\label{fig:dist_dataset}
\end{figure}

\begin{figure}[ht]
\centering
\includegraphics[width=\columnwidth]{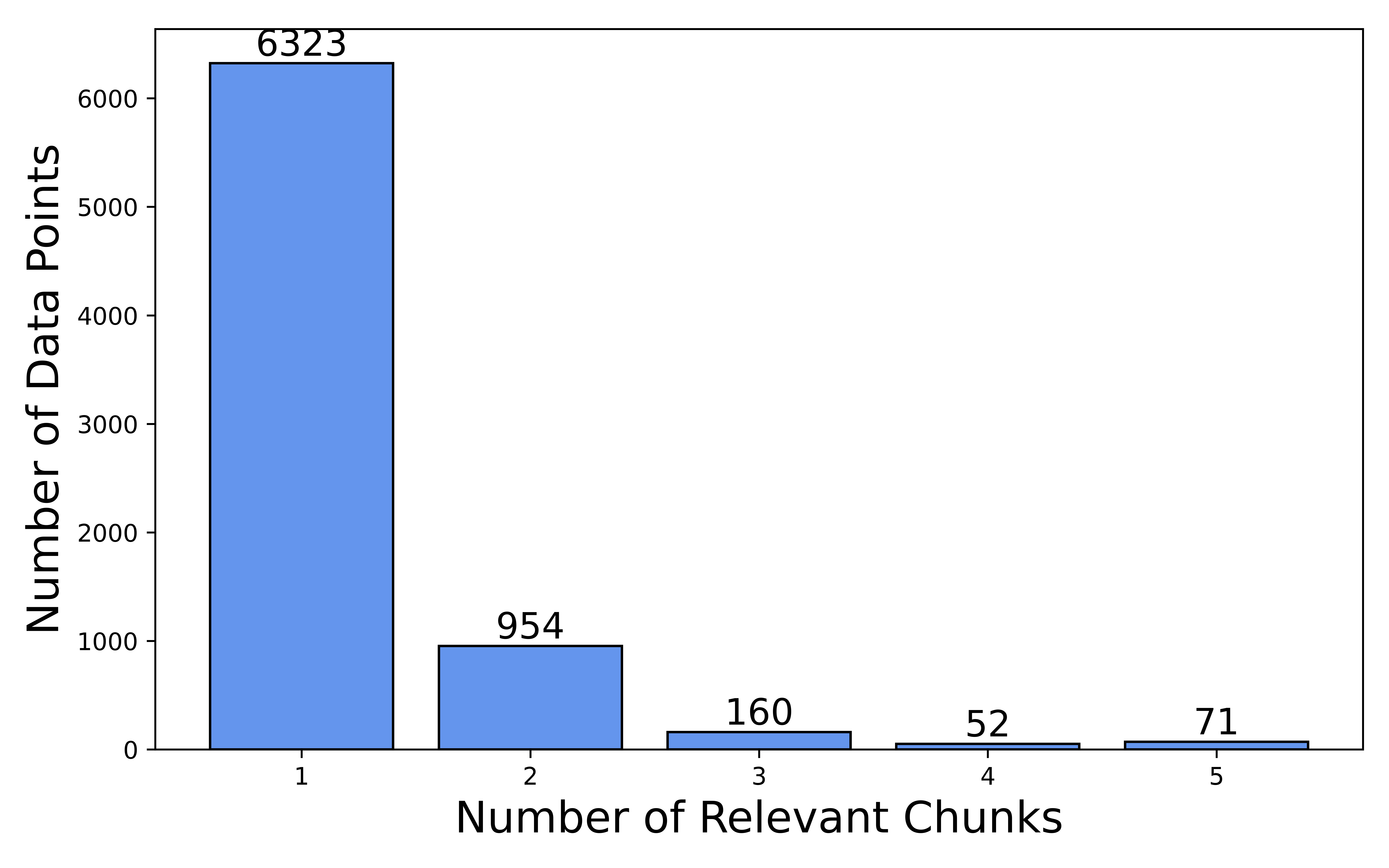}
\caption{Number of data points per relevant chunks}
\label{fig:dist_rel}
\end{figure}

Figure \ref{fig:dist_dataset} illustrates the number of data points derived from each source dataset within MiRAGE. The datasets include PopQA, NQ, IfQA, DROP, and TriviaQA.

\textbf{NQ} contributes the largest number of data points, totaling 3,578. This represents a significant portion of the overall dataset, reflecting NQ's comprehensive coverage.

\textbf{PopQA} provides 3,075 data points, slightly fewer than NQ, yet remains a substantial component of MiRAGE.

\textbf{TriviaQA} contributes 584 data points, offering a moderate addition to the dataset.

\textbf{IfQA} includes 248 data points, a smaller contribution indicating selective inclusion.

\textbf{DROP} offers the fewest with 75 data points, highlighting its more constrained role.

Figure \ref{fig:dist_rel} highlights the number of relevant chunks associated with each data point. The alignment of support and correctness labels defines the number of relevance. Albeit small in amount, the queries with 4 to 5 relevant chunks are cases where all of the chunks are from the reference article and the query can be inferred throughout the entire document. Such an example is shown in Table \ref{tab:data_sample}.

\input{Tables/tb_data_sample}

\section{Additional Experiments}
\label{sec:appendix_C}
\input{Tables/tb_rag_all}
Table \ref{tab:rag_all_top1}, \ref{tab:rag_all_top3}, and \ref{tab:rag_all_top5} provide detailed experimental results, encompassing four LLMs and five retrievers across three top-k settings. These tables collectively display a total of 60 configurations, highlighting the RAG adaptability of various RAG systems.

\section{Experimental Details}
We conducted all experiments using four RTX A6000 GPUs and utilized the vLLM framework to expedite inference \cite{kwon2023efficient}. In our work, we used GPT-4o (gpt-4o-2024-08-06) as a writing assistant. AI assistant was solely utilized for writing-related activities, such as grammar checking, refining awkward expressions, and translation of our manuscript. 

The models used in our experiments, along with their approximate parameter sizes, are listed below:

\begin{itemize}
    \item \textbf{BGE-S}: 33M parameters
    \item \textbf{BGE-B}: 110M parameters
    \item \textbf{BGE-L}: 335M parameters
    \item \textbf{Contriever}: 110M parameters
    \item \textbf{NV-embed-v2}: 7B parameters
    \item \textbf{E5-S}: 33M parameters
    \item \textbf{E5-B}: 110M parameters
    \item \textbf{E5-L}: 335M parameters
    \item \textbf{E5-Mistral}: 7B parameters
    \item \textbf{GTE-B}: 110M parameters
    \item \textbf{GTE-L}: 335M parameters
    \item \textbf{Llama2-7B}: 7B parameters
    \item \textbf{Llama2-70B}: 70B parameters
    \item \textbf{Llama3-8B}: 8B parameters
    \item \textbf{Qwen2-7B}: 7B parameters
\end{itemize}

\section{Human Validation Process}
\label{sec:appendix_E}

\begin{figure*}[t]
    \centering
    \includegraphics[width=1\linewidth]{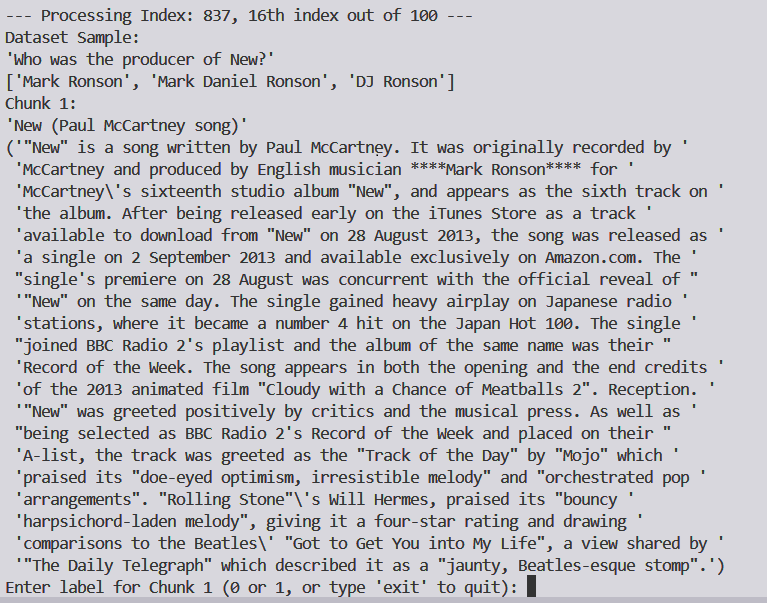}
    \caption{The command-line screen used for the annotation process.}
    \label{fig:annot}
\end{figure*}

For human validation, we employed three annotators, either native English speakers or those with a background in computational linguistics. Annotators were provided with detailed guidelines and underwent a short training phase to ensure consistency in labeling. Each annotator was given the query, answer spans, and the corresponding document chunk, including the title, as shown in Figure \ref{fig:annot}. For visual aid, answer spans were highlighted with '****' to draw attention to potentially relevant sections. However, annotators were instructed that the presence of an answer span does not directly indicate relevance.

Annotators A, B, and C each agreed with the model's labels for 467 (93.4\%), 477 (95.4\%), and 489 cases (97.8\%) out of 500, respectively. The pairwise inter-annotator agreement, measured using Cohen's Kappa, was 0.83 between A and B, 0.83 between A and C, and 0.89 between B and C, indicating strong agreement. The overall inter-annotator agreement, measured using Krippendorff's Alpha, was 0.8512, further validating the reliability of the annotation process. Annotators were fairly compensated for their efforts, and the process complied with standard ethical guidelines, including obtaining informed consent.

Figure \ref{fig:annot} illustrates the command-line annotation interface used in this process. The interface displays the query, potential answer spans, and the corresponding document chunk. Annotators could input their labels directly within the interface, ensuring an efficient and streamlined workflow.

\end{document}

%% file: Sections/1.Introduction.tex
\begin{figure*}[ht]
\centering
\includegraphics[width=\textwidth]{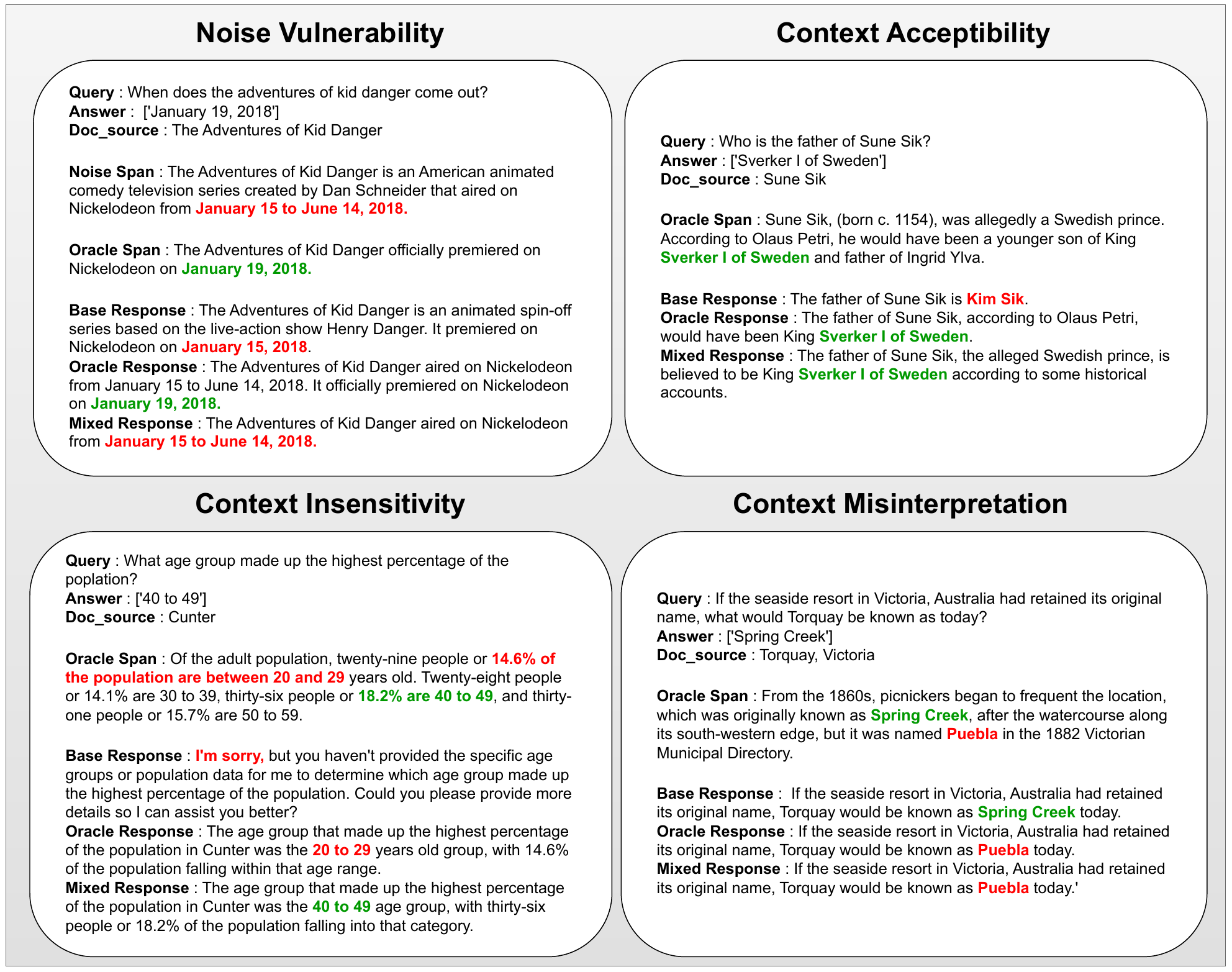}
\caption{Examples for four RAG Adaptability metrics. By analyzing model responses across three different settings, we assess the model's ability to utilize relevant information while disregarding irrelevant noise.}
\label{fig:metric}
\end{figure*}

Large Language Models (LLMs) have continuously advanced, demonstrating performance levels that increasingly surpass human capabilities \cite{achiam2023gpt, dubey2024llama}. Despite their expanding knowledge base, the capacity of parametric knowledge within LLMs is inherently limited \cite{yu2023generate, lewis2020retrieval}. As a result, LLMs face challenges when responding to information that emerges after their training period or when encountering data that is underrepresented within their training corpus \cite{mallen-etal-2023-trust, realtimeqa}.

To address these limitations, Retrieval-Augmented Generation (RAG) systems have been proposed as a practical solution \cite{Chameleon, gao2023retrieval, fan2024survey, hofstatter2023fid}. RAG enhances LLM performance by integrating external, non-parametric knowledge retrieved by a retrieval system, thereby extending the model’s ability to respond accurately beyond its parametric knowledge \cite{vu2023freshllms}. Research has demonstrated that RAG techniques improve domain adaptability \cite{hsieh2023tool} and mitigate hallucination issues \cite{ji2023towards}.

However, while RAG systems have advanced rapidly, research on robust and comprehensive evaluation methods lags behind. We identify several key challenges in the evaluation of RAG systems. First, the retrieval pools used for evaluation are often excessively large, making the process resource-intensive and inefficient \cite{mallen-etal-2023-trust, corr_survey}. For instance, many studies rely on the Wikipedia snapshot\footnote{\url{https://dumps.wikimedia.org/}}, containing over five million entries, for evaluating retrievers and RAG systems \cite{Chameleon, izacard-grave-2021-leveraging}. Indexing such large datasets incurs significant computational costs and introduces substantial delays. 

Second, current evaluation methods for RAG systems tend to focus disproportionately on performance improvements, often overlooking the complex dynamics between retrieval and generation \cite{xie2024adaptive, ru2024ragchecker}. Performance gains in generation are typically measured without considering critical aspects such as the effective integration of retrieved knowledge or knowledge conflicts within the system. 

Third, many LLM-based evaluation setups depend on large-scale external LLMs such as GPT-4 or Claude3, raising concerns regarding cost and accessibility \cite{es2023ragas}. This limits the scalability and replicability of evaluations across diverse research contexts.

To address these limitations, we introduce \textit{MIRAGE}, a compact yet challenging benchmark specifically designed for the evaluation of RAG systems. A lightweight proxy for computationally heavy RAG evaluation, MIRAGE comprises 7,560 queries linked to a retrieval pool of 37,800 document chunks. Each query is paired with at least one positive document chunk containing critical information for answering the question and several negative samples that are similar in content but lack key information. This setup enables precise and fast evaluation of both LLM and retriever performance while maintaining a smaller, more efficient retrieval pool. 

Furthermore, we propose four novel metrics that enable fine-grained analysis of both the generative performance of LLMs and their ability to integrate retrieved information. These metrics are specifically designed to evaluate RAG adaptability of a given LLM and retriever setup to find the optimal combination.

MIRAGE is crafted by reorganizing and refining existing benchmarks \cite{mallen2022not, kwiatkowski2019natural, joshi2017triviaqa, yu2023ifqa, dua2019drop}. In this paper, we fully demonstrate details of the data construction pipeline to ensure further reproducibility. We make our benchmark and code publicly accessible\footnote{Code and data will be released after publication}.

%% file: Sections/2.Related_Works.tex
Retrieval-Augmented Generation (RAG) has garnered significant attention in the field of natural language processing, leading to the development of various tools, benchmarks, and datasets aimed at evaluating system performance \cite{lewis2020retrieval, neelakantan2022text}. The current body of work primarily focuses on measuring the quality of retrieved context \cite{karpukhin2020dense}. However, existing solutions often have limitations, such as incomplete datasets or a lack of dedicated benchmarks that comprehensively cover both retrieval and generation tasks \cite{fabbri2021summeval}. This section reviews relevant tools, QA datasets, and benchmarks that have contributed to the evaluation of RAG systems, highlighting their strengths and areas where improvements are needed \cite{ yang2015wikiqa}.

\subsection{RAG Framework}
Recent advancements in Retrieval-Augmented Generation (RAG) have spurred the development of various evaluation tools and benchmarks \cite{gao2023retrieval}. However, existing solutions often suffer from limitations, either lacking comprehensive datasets or failing to sufficiently assess retriever performance. Several tools have emerged to evaluate RAG systems, focusing on metrics such as context relevance, answer faithfulness, and answer relevance. For example, RAGAS \cite{es2023ragas} provides a framework for evaluating these dimensions of RAG performance. Similarly, ARES \cite{saad2023ares} offers an automated evaluation system, utilizing lightweight language model judges fine-tuned on synthetic data to assess both retrieval and generation components. Additionally, RAGCHECKER \cite{ru2024ragchecker} enables detailed analysis of retrieval and generation within RAG systems. While these tools offer valuable insights through diverse evaluation metrics, they often lack dedicated datasets tailored for benchmarking RAG performance comprehensively.

\subsection{QA Datasets}
Several Question Answering (QA) datasets have been developed to challenge Large Language Models (LLMs) with queries that are difficult to answer without relevant context. Examples include PopQA, TriviaQA, IfQA, and DROP, which are primarily based on Wikipedia data and are designed to expose performance variations in RAG settings. TriviaQA \cite{joshi2017triviaqa}, for instance, contains over 650K question-answer-evidence triples derived from 95K trivia enthusiast-authored pairs, with an average of six supporting evidence documents per question. Similarly, PopQA \cite{mallen2022not} is a large-scale open-domain QA dataset comprising 14K entity-centric question-answer pairs. Although these datasets provide valuable QA pairs, they lack integrated retrieval pools, requiring researchers to develop their own retrieval systems and data-loading processes, which can complicate experimentation and limit reproducibility.

\subsection{RAG Benchmarks}
A few benchmarks, such as RGB \cite{chen2024benchmarking} and RECALL \cite{liu2023recall}, provide datasets specifically designed for RAG evaluation. Despite their contributions, these benchmarks often fall short in thoroughly assessing retriever performance, which is a critical component in RAG systems. Large-scale datasets like Natural Questions (NQ) \cite{kwiatkowski2019natural} and MS MARCO \cite{bajaj2016ms} have been widely adopted in information retrieval and question-answering tasks, maintaining leaderboards and benchmarks for broader community use. However, these datasets rely on entire Wikipedia dumps, which, while comprehensive, are impractically large for local retrieval pool construction. For example, the NQ corpus requires QA systems to process entire Wikipedia articles, many of which may not contain the relevant answers, leading to inefficiencies in both retrieval and evaluation.

To the best of our knowledge, there is currently no publicly available benchmark that provides both question-answer pairs and corresponding retrieval pools designed specifically for RAG system evaluation. This gap underscores the need for a comprehensive benchmark that facilitates both the assessment of RAG systems and the provision of easily accessible retrieval pools, enabling more efficient and reproducible experimentation.

%% file: Sections/3.MiRAGE.tex
The MIRAGE dataset is designed as a high-quality benchmark aimed at evaluating the diverse components of RAG systems through a challenging set of question-answer pairs. To ensure robustness and relevance, we employed a meticulous multi-stage filtering process, as illustrated in Figure \ref{fig:filtering}. Below, we outline each stage of dataset construction, from initial selection to the multi-layered filtering that guarantees data quality.

\subsection{Dataset Selection}
We initiated the construction of MIRAGE by selecting existing QA datasets that satisfy three primary criteria: (1) Wikipedia-based content, (2) availability of answer spans, and (3) the inclusion of document information. Datasets such as PopQA, Natural Questions (NQ), TriviaQA, IfQA, and DROP were chosen due to their alignment with these criteria. These datasets either provide document titles directly (PopQA, NQ) or contain passages traceable to full Wikipedia articles (TriviaQA, IfQA, DROP). Datasets focusing on multi-hop retrieval, such as HotpotQA or WikiHop \cite{yang2018hotpotqa, li2021hopretriever}, were excluded to concentrate on single-hop retrieval scenarios. This selection process resulted in an initial pool of over 500,000 QA pairs from five distinct datasets.

\subsection{Query-to-Document Mapping}
For query-document alignment, we collected over 6 million Wikipedia articles using the enwiki dump from September 2024\footnote{We use the enwiki dump 20240901 from \url{https://dumps.wikimedia.org/enwiki/20240901/}}. In contrast to standard dataset construction practices where queries are generated from documents, we reversed the process by mapping existing queries back to their respective Wikipedia articles. Using Elasticsearch, we processed datasets with fragmented document presentations to efficiently link queries with articles. During this phase, we filtered out unmappable queries and those with duplicate document sources to streamline the dataset and ensure diverse topic coverage. This step resulted in a refined set of 61,165 QA pairs accurately mapped to Wikipedia articles.

\subsection{Document Chunking}
To facilitate efficient retrieval, we segmented the Wikipedia articles into 330-token chunks using the BERT-base-uncased tokenizer. This segmentation employed a recursive, sentence-level strategy to preserve sentence integrity while minimizing information loss. The 330-token length was determined to offer an optimal balance, as preliminary experiments indicated that this chunk size outperforms alternatives (e.g., 110 or 550 tokens) in retrieving both relevant and negative samples. This process generated a total of 16,508,989 document chunks, each indexed by document title for efficient querying. For each query, we retrieved the top 5 document chunks by title match, assuming that answers would likely reside within the corresponding documents. This assumption was rigorously tested through subsequent filtering steps.

\subsection{Multi-Layered Filtering}
To ensure the quality and challenge level of the dataset, we applied a multi-layered filtering methodology. This filtering process focused on refining both positive and negative samples to guarantee the reliability of retrieval and generation evaluations. 

\paragraph{Step 1: Support Labeling}
\label{sec:sup}
Given the high cost of manual relevance judgments, we utilized the C4AI command-r model \cite{cohere2024command}, specifically optimized for RAG systems, to assign support labels. The model evaluated whether the retrieved chunks provided sufficient context to answer the given query. This automated step approximated human judgment in assessing chunk relevance and significantly reduced manual annotation costs. The detailed prompt for this task is shown in Appendix \ref{sec:appendix_A}.

\paragraph{Step 2: Validation through Inference}
\label{sec:val}
To further validate chunk relevance, we employed Llama-3.1-8B Instruct \cite{dubey2024llama} for inference-based validation. The model was tasked with answering queries using the retrieved chunks, and instances where the model accurately responded were labeled as valid. Additionally, we filtered out instances where the model could answer the query without requiring the provided context, ensuring that the remaining data points present a higher level of difficulty for RAG evaluation.

\paragraph{Step 3: Document Title Verification}
For chunks that passed the previous validation steps, we ensured that at least one relevant chunk per query was accurately mapped to the query’s reference document. This step was critical to confirming that the chunk contained the correct answer from the original dataset and that the model was not relying on irrelevant information to generate the response.

\subsection{Human Validation}
Since the aforementioned filtering process heavily relies on the reasoning capabilities of large language models , the quality of the automatically generated labels is not guaranteed. To validate these labels, we conducted human validation by randomly sampling 100 queries and 500 document chunks mapped to each query. Three annotators were asked to label whether the query could be answered based on the information provided in the chunk. 

On average, the annotators exhibited 95\% agreement with the model's labels, demonstrating a strong alignment with the automatic labeling process. The inter-annotator agreement, measured using Krippendorff's alpha, was 0.85, indicating substantial agreement among the annotators. Further, pairwise, Cohen's Kappa scores ranged from 0.83 to 0.89 for the three annotators, reinforcing the reliability of the annotations. 

Details of the annotation process are shown in Appendix \ref{sec:appendix_E}, and an example of the annotation interface is illustrated in Figure \ref{fig:annot}.

\subsection{Final Dataset Statistics}
The finalized MIRAGE consists of 7,560 QA pairs mapped to a retrieval pool of 37,800 document chunks. Each query is associated with one or more positive chunks and several negative samples, enabling precise evaluation of retrievers, LLMs, and RAG systems. This structured dataset facilitates efficient, fine-grained analysis of retrieval and generation components in a RAG setting. Further details of the prompt design and filtering methodologies are provided in Appendix \ref{sec:appendix_A}.

%% file: Sections/4.Evaluation_Measure.tex
To thoroughly assess the performance of RAG systems, we define three distinct evaluation setups: a base response without context, an oracle response with the correct context, and a mixed response containing both noisy and oracle chunks. The mixed response setup mirrors real-world RAG settings since, in practice, RAG systems process both noisy and relevant information simultaneously. In contrast, the base and oracle settings serve as the lower and upper performance bounds, respectively, for each system. We consistently observe that a model’s performance falls between its base and oracle performance in every scenario. By analyzing model behavior across these setups, we identify the system's strengths and vulnerabilities in handling external knowledge.

\subsection{Input Context Configurations}

We evaluate the LLM and retrieval components of RAG systems under three distinct input settings. Performance is measured using exact match accuracy between the system’s output and the correct answer label, ensuring a rigorous assessment of both retriever and LLM components\footnote{MIRAGE supports evaluation of various setups including LLM-only, retriever-only, and LLM with retriever, enabling flexible framework for various use cases.}.

\paragraph{Base Setting ($\mathbf{Ans}_{\textbf{B}}$):} In this configuration, the LLM generates an answer based solely on its internal parametric knowledge, with no external context. This serves as a baseline for evaluating the inherent knowledge embedded in the LLM without augmentation from retrieval.

\paragraph{Oracle Context Setting ($\mathbf{Ans}_{\textbf{O}}$):} In this setup, the LLM is provided with only the correct context chunk, free of noise or irrelevant information. One relevant chunk is selected from the top-5 chunks mapped to each query. This setup evaluates the LLM’s ability to deliver accurate answers when supplied with highly relevant information.

\paragraph{Mixed Context Setting ($\mathbf{Ans}_{\textbf{M}}$):} Here, the LLM receives a mixture of five chunks, including one relevant (oracle) chunk and several irrelevant (noise) chunks. The distribution of relevant chunks per query is shown in Figure \ref{fig:dist_rel}. This setup tests the model's robustness in differentiating between relevant and irrelevant information within a noisy retrieval context.

By comparing the LLM’s performance across these three settings, we assess its intrinsic knowledge capabilities, its ability to leverage external context, and its robustness against noisy information. We determine accuracy by checking for an exact match between the output of the generated RAG system and the label word or sentence. The output for each query, denoted as $\mathbf{Ans}$, receives a binary score based on the exact match label.

\subsection{RAG Adaptability Metrics}

With the notion of our pre-defined three cases, we define a subgroup $G(b, o, m) \subset D$ as Equation~\ref{eq:subgroup}, where $D$ is a whole dataset. Here, $b$, $o$, and $m$ are binary variables taking values of either 0 or 1. They serve as variables to define groups corresponding to each case based on the generation results of the RAG system. 
    
\begin{equation} \label{eq:subgroup}
\begin{split}
    G(b, o, m) = \{d \in D \,|\, \mathbf{Ans}_{\textbf{B}}(d) = b \, \wedge \\
    \mathbf{Ans}_{\textbf{O}}(d) = o \,\wedge\, \mathbf{Ans}_{\textbf{M}}(d) = m \}
\end{split}
\end{equation}

Using this indicator, we introduce four novel metrics designed to capture the nuanced interactions between the retrieval and generation components of RAG systems. These metrics provide a detailed analysis of model behavior under various input conditions, revealing the system's adaptability and potential weaknesses. Detailed examples are provided in Figure \ref{fig:metric}.

\paragraph{Noise Vulnerability:} 
This metric assesses the model’s susceptibility to noise in the context. Specifically, it captures instances where the model provides incorrect answers due to irrelevant information, even when the correct context is present. These cases occur when the model fails under the mixed context ($\mathbf{Ans}_{\textbf{M}}(d)=0$), but succeeds when given the oracle context ($\mathbf{Ans}_{\textbf{O}}(d)=1$), indicating difficulty in filtering out irrelevant chunks.
\begin{equation}
    \cfrac{|G(0, 1, 0)| + |G(1, 1, 0)|}{|D|}
\end{equation}

\paragraph{Context Acceptability:}
This metric evaluates the model’s ability to effectively leverage the provided context to generate accurate answers. It captures scenarios where the model answers correctly in both the oracle and mixed contexts ($\mathbf{Ans}_{\textbf{M}}(d)=\mathbf{Ans}_{\textbf{O}}(d)=1$), indicating robustness in processing noisy inputs while accurately extracting relevant information.
\begin{equation}
    \cfrac{|G(0, 1, 1)| + |G(1, 1, 1)|}{|D|}
\end{equation}

\paragraph{Context Insensitivity:}
This metric highlights cases where the model fails to utilize the context information, producing incorrect answers regardless of whether the correct context is provided. Specifically, it tracks instances where the model's base and oracle responses are both incorrect ($\mathbf{Ans}_{\textbf{B}}=\mathbf{Ans}_{\textbf{O}}=0$), revealing challenges in integrating external knowledge into its reasoning process.
\begin{equation}
    \cfrac{|G(0, 0, 0)| + |G(0, 0, 1)|}{|D|}
\end{equation}

\input{Tables/tb_rag}

\paragraph{Context Misinterpretation:}
A hallucination occurs when the model generates incorrect responses even with the correct context provided. This metric identifies cases where the model answers correctly without context ($\mathbf{Ans}_{\textbf{B}}=1$) but produces incorrect answers when given the oracle context ($\mathbf{Ans}_{\textbf{O}}=0$). Such instances indicate that the model is either misinterpreting the context or over-relying on irrelevant information, leading to hallucinated outputs.
\begin{equation}
    \cfrac{|G(1, 0, 0)| + |G(1, 0, 1)|}{|D|}
\end{equation}

\subsection{Comprehensive RAG Evaluation}

Since the four metrics cover all possible cases, they add up to 1, enabling a system-level analysis and revealing an LLM's possible weaknesses and strengths. This framework not only highlights the retrieval system's role in enhancing or hindering generative performance but also provides critical insights into the model's behavior in real-world scenarios where relevant and irrelevant information are mixed. Figure \ref{fig:metric} illustrates examples for each category, providing a visual representation of the different error patterns captured by our evaluation methodology.

\begin{equation}
    \sum_{b,o,m \in \{0,1\}}\cfrac{|G(b, o, m)|}{|D|} = 1
\end{equation}

%% file: Tables/tb_rag.tex
\definecolor{metacolorab}{HTML}{F5F5F7}

\begin{table*}[ht]
\centering
\resizebox{0.8\textwidth}{!}{
\begin{tabular}{c|c|cccc}
\toprule[1.5pt]

\makecell[c]{\textbf{LLM}} & \makecell[c]{\textbf{Retriever}} & \makecell[c]{\textbf{Noise} \\ \textbf{Vulnerability}($\downarrow$)} &
\makecell[c]{\textbf{Context} \\ \textbf{Acceptability} ($\uparrow$)} &
\makecell[c]{\textbf{Context} \\ \textbf{Insensitivity} ($\downarrow$)} &
\makecell[c]{\textbf{Context} \\ \textbf{Misinterpretation} ($\downarrow$)} \\ \midrule[1.5pt]

\rowcolor{metacolorab} \multicolumn{6}{c}{\textit{\textbf{Top-1}}} \\
\midrule[1.5pt]

\multirow{3}{*}{\makecell[l]{\textbf{LLAMA2-7B}}}
& \textbf{Contriever} & 47.63 & 35.33 & 16.37 & 0.67 \\
& \textbf{BGE-Base} & 25.77 & 57.19 & 16.37 & 0.67 \\
& \textbf{nv-embed-v2} & \textbf{18.21} & \textbf{64.75} & 16.37 & 0.67 \\
\midrule
\multirow{3}{*}{\makecell[l]{\textbf{GPT3.5}}}
& \textbf{Contriever} & 42.26 & 48.76 & 8.44 & 0.54 \\
& \textbf{BGE-Base} & 24.36 & 66.67 & 8.44 & 0.54 \\
& \textbf{nv-embed-v2} & \textbf{17.67} & \textbf{73.36} & 8.44 & 0.54 \\
\midrule
\multirow{3}{*}{\makecell[l]{\textbf{GPT-4o}}}
& \textbf{Contriever} & 35.73 & 55.42 & 8.29 & 0.57 \\
& \textbf{BGE-Base} & 20.57 & 70.57 & 8.29 & 0.57 \\
& \textbf{nv-embed-v2} & \textbf{15.38} & \textbf{75.76} & 8.29 & 0.57 \\

\midrule[1.5pt]
\rowcolor{metacolorab} \multicolumn{6}{c}{\textit{\textbf{Top-3}}} \\
\midrule[1.5pt]

\multirow{3}{*}{\textbf{LLAMA2-7B}} 
& \textbf{Contriever} & 36.56 & 46.40 & 16.37 & 0.67 \\
& \textbf{BGE-Base} & 21.72 & 61.23 & 16.37 & 0.67 \\
& \textbf{nv-embed-v2} & \textbf{16.81} & \textbf{66.15} & 16.37 & 0.67 \\
\midrule
\multirow{3}{*}{\textbf{GPT3.5}}
& \textbf{Contriever} & 30.91 & 60.11 & 8.44 & 0.54 \\
& \textbf{BGE-Base} & 17.39 & 73.64 & 8.44 & 0.54 \\
& \textbf{nv-embed-v2} & \textbf{13.16} & \textbf{77.87} & 8.44 & 0.54 \\
\midrule
\multirow{3}{*}{\textbf{GPT-4o}}
& \textbf{Contriever} & 25.94 & 65.20 & 8.29 & 0.57 \\
& \textbf{BGE-Base} & 13.67 & 77.47 & 8.29 & 0.57 \\
& \textbf{nv-embed-v2} & \textbf{10.75} & \textbf{80.39} & 8.29 & 0.57 \\

\midrule[1.5pt]
\rowcolor{metacolorab} \multicolumn{6}{c}{\textit{\textbf{Top-5}}} \\
\midrule[1.5pt]

\multirow{3}{*}{\textbf{LLAMA2-7B}} 
& \textbf{Contriever} & 36.78 & 46.17 & 16.37 & 0.67 \\
& \textbf{BGE-Base} & 24.08 & 58.88 & 16.37 & 0.67 \\
& \textbf{nv-embed-v2} & \textbf{19.93} & \textbf{63.03} & 16.37 & 0.67 \\
\midrule
\multirow{3}{*}{\textbf{GPT3.5}}
& \textbf{Contriever} & 27.45 & 63.57 & 8.44 & 0.54 \\
& \textbf{BGE-Base} & 16.42 & 74.60 & 8.44 & 0.54 \\
& \textbf{nv-embed-v2} & \textbf{13.21} & \textbf{77.81} & 8.44 & 0.54 \\
\midrule
\multirow{3}{*}{\textbf{GPT-4o}}
& \textbf{Contriever} & 22.65 & 68.49 & 8.29 & 0.57 \\
& \textbf{BGE-Base} & 12.79 & 78.36 & 8.29 & 0.57 \\
& \textbf{nv-embed-v2} & \textbf{10.64} & \textbf{80.50} & 8.29 & 0.57 \\

\bottomrule[1.5pt]
\end{tabular}
}
\caption{RAG adaptability scores for various RAG systems. We present representative performance results for combinations involving three different LLMs, three different retrievers, and three top-k setups, totaling 27 model combinations. A comprehensive experiment covering all 60 configurations is detailed in Appendix \ref{sec:appendix_C}}
\label{tab:rag_adaptability_main}
\end{table*}

%% file: Sections/5.Experiments.tex
This section presents the experimental setup, including the dataset, models, and evaluation metrics used in our study, followed by a comprehensive analysis of the experimental results.

\subsection{Dataset}

The MIRAGE dataset is designed to evaluate RAG systems across a range of question-answering tasks. It comprises 7,560 QA pairs, each mapped to a retrieval pool of 37,800 document chunks. For each query, we include a mix of relevant and irrelevant document chunks to test the system's ability to filter out noise and identify the correct information. The dataset is balanced across different domains and contexts, ensuring a comprehensive assessment of retrieval and generation capabilities.

\subsection{Models}

We evaluated a combination of retrievers and LLMs in RAG settings. The retrievers used include models of various sizes and architectures, such as BGE \cite{chen2024bge}, E5 \cite{wang2024multilingual}, Contriever \cite{izacard2021unsupervised}, GTE \cite{zhang2024mgte}, and nv-embed-v2 \cite{lee2024nv}. For the generation, we used five LLMs: Llama-2-7B-Chat, Llama-2-70B-Chat \cite{touvron2023llama}, GPT-3.5-Turbo, GPT-4o, and QWEN2-7B-Instruct \cite{yang2024qwen2}. These LLMs represent a diverse range of performance levels, from moderately sized models to state-of-the-art systems.




\subsection{Results and Analysis}

\paragraph{RAG System Performance:}  
The results, summarized in Table \ref{tab:rag_adaptability_main}, show that RAG systems exhibit varying levels of performance depending on the number of retrieved chunks and the quality of both the retriever and the LLM. Generally, increasing the number of retrieved chunks from Top-1 to Top-3 enhances performance due to the higher likelihood of including the oracle chunk. However, advancing to Top-5 retrieval often introduces additional noise, leading to performance degradation in some configurations. This effect is particularly prominent in models like GPT-3.5-Turbo and Llama-2-7B-Chat, which show decreased scores when handling additional noisy chunks.

The combination of GPT-4o and nv-embed-v2 show robust performance across all retrieval settings, maintaining high scores even with the introduction of additional chunks. This indicates the model's strong capacity for filtering out irrelevant information and focusing on the relevant chunks. In contrast, the performance of Llama-2 models was more sensitive to noise, suggesting that these models benefit less from additional context when the retrieval results contain irrelevant information.

Moreover, whereas noise vulnerability and context acceptability drastically change with retriever performance, cases where Oracle information is not utilized—namely, context insensitivity and context misinterpretation—are consistent with each model regardless of given shots or retrievers. This indicates that the ability to utilize the context properly relies solely on the LLM's capabilities. Consequently, this reliance explains why overall performance does not achieve perfect scores in Oracle settings. Although these metrics do not distinguish between retrievers, they provide valuable insights into the LLM's weaknesses.

\input{Tables/tb_retriever}

\paragraph{Retriever Performance:}
Comprising 37,800 distinctive document chunks, MIRAGE is also a valuable tool for evaluating retriever performance. Table \ref{tab:retrieval_performance} presents the performance of various retrieval models in terms of F1 and NDCG scores. The results demonstrate that the MIRAGE benchmark effectively differentiates performance across different model sizes and architectures. Larger and more recent models, such as nv-embed-v2, consistently outperform smaller retrievers like BGE and Contriever. These results align with previous studies, which demonstrate that more advanced retrieval architectures lead to enhanced retrieval accuracy. Notably, the nv-embed-v2 model consistently retrieves more relevant chunks, resulting in higher overall RAG system performance, particularly when paired with high-performing LLMs.

\paragraph{LLM Performance:}
The performance of LLMs with fixed context setting is reported in Table \ref{tab:benchmark_data}. In this setup, we give an LLM a fixed set of 5 document chunks mapped to each query. This setup enables a swift assessment of various LLMs' capabilities without relying on retrievers and solely using the MIRAGE dataset. 

Both GPT-4o and GPT-3.5-Turbo exhibit the highest accuracy, demonstrating strong abilities in answering questions without external context. Although Llama-2-7B-Chat and QWEN2-7B-Instruct initially show lower performance, they exhibit significant improvement when integrated with retrieval systems, highlighting the positive impact of retrieval augmentation on weaker models. These results illustrate that retrieval can effectively help bridge the performance gap between smaller models and state-of-the-art systems.

\input{Tables/tb_llm_ideal}



Overall, the results demonstrate MIRAGE's ability to offer a nuanced evaluation of RAG systems through a robust metrics-driven framework. This approach uncovers both the strengths and weaknesses of different model combinations across various levels of retrieval context.

%% file: Tables/tb_retriever.tex
\definecolor{metacolorab}{HTML}{F5F5F7}

\begin{table}[t]
\centering
\small
\begin{tabular}{l|cccc}
\toprule[1.5pt]
\makecell[c]{\textbf{Model}} & \makecell[c]{\textbf{F1}} & \makecell[c]{\textbf{Precision}} & \makecell[c]{\textbf{Recall}} & \makecell[c]{\textbf{NDCG}} \\

\midrule[1.5pt]
\rowcolor{metacolorab} \multicolumn{5}{c}{\textit{\textbf{Top-1}}} \\
\midrule[1.5pt]

\textbf{BGE-S} & 63.03 & 67.87 & 60.99 & 67.87 \\
\textbf{BGE-B} & 64.12 & 68.94 & 62.08 & 68.94 \\
\textbf{BGE-L} & 68.60 & 73.73 & 66.43 & 73.73 \\
\textbf{E5-S} & 64.32 & 68.97 & 62.35 & 68.97 \\
\textbf{E5-B} & 63.54 & 68.13 & 61.61 & 68.13 \\
\textbf{E5-L} & 71.38 & 76.65 & 69.14 & 76.65 \\
\textbf{GTE-B} & 59.40 & 63.94 & 57.50 & 63.94 \\
\textbf{GTE-L} & 63.29 & 67.98 & 61.33 & 67.98 \\
\textbf{Contriever} & 39.82 & 43.25 & 38.40 & 43.25 \\
\textbf{E5-Mistral} & 67.96 & 73.07 & 65.81 & 73.07 \\
\textbf{NV} & \textbf{73.92} & \textbf{79.40} & \textbf{71.60} & \textbf{79.40} \\

\midrule[1.5pt]
\rowcolor{metacolorab} \multicolumn{5}{c}{\textit{\textbf{Top-3}}} \\
\midrule[1.5pt]

\textbf{BGE-S} & 45.31 & 32.35 & 82.95 & 78.34 \\
\textbf{BGE-B} & 45.82 & 32.70 & 83.95 & 79.42 \\
\textbf{BGE-L} & 47.71 & 34.09 & 87.24 & 83.03 \\
\textbf{E5-S} & 45.00 & 31.95 & 82.93 & 78.92 \\
\textbf{E5-B} & 45.59 & 32.47 & 83.74 & 78.76 \\
\textbf{E5-L} & 48.47 & 34.53 & 88.93 & 85.55 \\
\textbf{GTE-B} & 44.32 & 31.64 & 81.20 & 75.58 \\
\textbf{GTE-L} & 46.42 & 33.14 & 84.99 & 79.36 \\
\textbf{Contriever} & 34.38 & 24.61 & 62.84 & 56.17 \\
\textbf{E5-Mistral} & 47.61 & 33.82 & 87.71 & 83.34 \\
\textbf{NV} & \textbf{50.77} & \textbf{36.35} & \textbf{92.56} & \textbf{88.05} \\

\midrule[1.5pt]
\rowcolor{metacolorab} \multicolumn{5}{c}{\textit{\textbf{Top-5}}} \\
\midrule[1.5pt]

\textbf{BGE-S} & 32.98 & 20.92 & 88.12 & 80.70 \\
\textbf{BGE-B} & 33.42 & 21.21 & 89.25 & 81.84 \\
\textbf{BGE-L} & 34.51 & 21.92 & 92.02 & 85.20 \\
\textbf{E5-S} & 32.76 & 20.69 & 88.26 & 81.41 \\
\textbf{E5-B} & 33.39 & 21.15 & 89.46 & 81.40 \\
\textbf{E5-L} & 34.69 & 21.97 & 92.97 & 87.40 \\
\textbf{GTE-B} & 32.66 & 20.72 & 87.31 & 78.37 \\
\textbf{GTE-L} & 33.84 & 21.49 & 90.32 & 81.76 \\
\textbf{Contriever} & 26.78 & 17.02 & 71.43 & 60.00 \\
\textbf{E5-Mistral} & 34.36 & 21.72 & 92.40 & 85.51 \\
\textbf{NV} & \textbf{36.38} & \textbf{23.18} & \textbf{96.41} & \textbf{89.78} \\

\bottomrule[1.5pt]

\end{tabular}
\caption{Performance comparison of various retrieval models on MIRAGE dataset. S, B, and L denote Small, Base, and Large model variants respectively. Bold indicates the best performance for each metric and Top-k setting.}
\label{tab:retrieval_performance}
\end{table}

%% file: Tables/tb_llm_ideal.tex
\begin{table}[H] 
\centering
\small
\begin{tabular}{l|c|c|c}

\toprule[1.5pt]
\makecell[c]{\textbf{Models}} & \makecell[c]{\textbf{Base}} & \makecell[c]{\textbf{Mixed} \\ \textbf{Context}} & 
\makecell[c]{\textbf{Oracle} \\ \textbf{Context}} \\ \midrule[1.5pt]

\textbf{LLAMA2-7B} & 6.60 & 74.19 & 82.96 \\
\textbf{LLAMA2-70B} & 15.75 & 80.33 & 87.57 \\
\textbf{Qwen2-7B} & 7.39 & 83.74 & 90.22 \\
\textbf{GPT3.5} & 31.96 & 87.27 & 91.02 \\
\textbf{GPT4o} & \textbf{45.82} & \textbf{87.49} & \textbf{91.14} \\

\bottomrule[1.5pt]

\end{tabular}
\caption{Performance comparison of various LLMs on the MIRAGE dataset: The base setting evaluates the LLM's internal knowledge without any context. The mixed context setting assesses the LLM's ability to utilize relevant chunks while disregarding irrelevant information. The oracle context tests whether the LLM can effectively employ necessary information for accurate inference. Scores are reported as accuracy.}

\label{tab:benchmark_data}
\end{table}

%% file: Tables/tb_prompt_base.tex
\definecolor{metacolorscore}{HTML}{FAF7F0}

\begin{table}[h]
\centering
\resizebox{0.8\linewidth}{!}{
\begin{tabular}{l}

\toprule[1.5pt]
\rowcolor{metacolorscore} \makecell[l]{
\textbf{System Prompt} \\
You are a helpful assistant. \\
\midrule
\textbf{User Prompt} \\
\textcolor{brown}{\textbf{Question}} : What is John Mayne\'s occupation?               \\
\textcolor{brown}{\textbf{Answer}} :                                                   \\
\midrule
\textbf{Model Response} \\
\textcolor{brown}{\textbf{I'm sorry but I have no information ...}} \\
} \\
\bottomrule[1.5pt]

\end{tabular}} \caption{Inference prompt for the base setup.} \label{tab:base_prompt}
\end{table}

%% file: Tables/tb_prompt_oracle.tex
\definecolor{metacolorscore}{HTML}{FAF7F0}

\begin{table}[h]
\centering
\resizebox{0.97\linewidth}{!}{
\begin{tabular}{l}

\toprule[1.5pt]
\rowcolor{metacolorscore} \makecell[l]{
\textbf{System Prompt} \\
You are a helpful assistant. \\
\midrule
\textbf{User Prompt} \\
\textcolor{brown}{\textbf{Question}} : What is John Mayne\'s occupation? \\
\textcolor{brown}{\textbf{Context}} : Scottish printer, journalist and \\
poet John Mayne (1759–1836) was a Scottish printer, \\
journalist and poet born in Dumfries. In 1780, his \\
poem "The Siller Gun" appeared in its original form \\
in "Ruddiman\'s Magazine", published by Walter \\
Ruddiman in Edinburgh. It is a humorous work on an \\
ancient custom in Dumfries of shooting for the \\
"Siller Gun." He also wrote a poem on "Halloween" \\
in 1780 which influenced Robert Burns\'s 1785 poem \\
"Halloween". Mayne also wrote a version of the \\
ballad "Helen of Kirkconnel". His verses were \\
admired by Walter Scott. Life. He was born at Dumfries \\
on 26 March 1759. Educated at the local grammar \\
school, he became a printer in the office of the \\
"Dumfries Journal". In 1782 he went with his family \\
to Glasgow, where he worked for five years in the \\
publishing house of the brothers Foulis. In 1787 \\
he settled in London, first as a printer, and then \\
as proprietor and joint editor of "The Star", an \\
evening paper, in which he placed his poems. He \\
died at Lisson Grove, London, 14 March 1836. Works. \\
Mayne wrote poetry in Dumfries, and after 1777 he \\
contributed poems to "Ruddiman\'s Weekly Magazine", \\
Edinburgh. Between 1807 and 1817 several of his \\
lyrics appeared in the "Gentleman\'s Magazine \\
\textcolor{brown}{\textbf{Answer}} : \\
\midrule
\textbf{Model Response} \\
\textcolor{brown}{\textbf{Printer, journalist, and poet}} \\
} \\
\bottomrule[1.5pt]

\end{tabular}} \caption{Inference prompt for the oracle setup.} \label{tab:oracle_prompt}
\end{table}

%% file: Tables/tb_prompt_mixed.tex
\definecolor{metacolorscore}{HTML}{FAF7F0}

\begin{table}[h]
\centering
\resizebox{0.97\linewidth}{!}{
\begin{tabular}{l}

\toprule[1.5pt]
\rowcolor{metacolorscore} \makecell[l]{
\textbf{System Prompt} \\
You are a helpful assistant. \\
\midrule
\textbf{User Prompt} \\
\textcolor{brown}{\textbf{Question}} : What is John Mayne\'s occupation? \\
\textcolor{brown}{\textbf{Context}} : \\
1. Scottish printer, journalist and poet John \\
Mayne (1759–1836) was a Scottish printer, \\
journalist and poet born in... \\
2. Mayne\'s "Siller Gun" was based on a \\
Dumfries wapinschaw: the competitors were \\
members of the corporations, and the prize... \\
3.British lawyer (1828–1917) John Dawson Mayne \\
(1828–1917) was a British lawyer and legal expert \\
who served as acting Advocate-General... \\
4. Mayne served as the Professor of law, logic \\
and moral philosophy at the Presidency College, \\
Madras from 1857 throughout the 1860s. He also... \\
5. Annie\'s first husband\'s name is unknown, \\
but she was the daughter of Charles Craigie-  \\
Halkett-Inglis of Hallhill, Fife and Cramond...  \\ 
\textcolor{brown}{\textbf{Answer}} : \\
\midrule
\textbf{Model Response} \\
\textcolor{brown}{\textbf{British Lawyer}} \\
} \\
\bottomrule[1.5pt]

\end{tabular}} \caption{Inference prompt for the mixed setup.} \label{tab:mixed_prompt}
\end{table}

%% file: Tables/tb_prompt_command.tex
\definecolor{metacolorscore}{HTML}{FAF7F0}

\begin{table}[h]
\centering
\resizebox{0.97\linewidth}{!}{
\begin{tabular}{l}

\toprule[1.5pt]
\rowcolor{metacolorscore} \makecell[l]{
\textbf{System Prompt} \\
You are an accurate and reliable AI assistant \\
that can answer questions with the help of external \\
documents. Please note that external documents may \\
contain noisy or factually incorrect information. \\
If the information in the document contains the \\
correct answer, you will generate 'Supported'. If \\
the information in the document does not contain the \\
answer, you will generate 'Not supported.' \\
\midrule
\textbf{User Prompt} \\
\textcolor{brown}{\textbf{Document}} : Scottish printer, journalist and \\
poet John Mayne (1759–1836) was a Scottish printer, \\
journalist and poet born in Dumfries. In 1780, his \\
poem "The Siller Gun" appeared in its original form \\
in "Ruddiman\'s Magazine", published by Walter \\
Ruddiman in Edinburgh. It is a humorous work on an \\
ancient custom in Dumfries of shooting for the \\
"Siller Gun." He also wrote a poem on "Halloween" \\
in 1780 which influenced Robert Burns\'s 1785 poem \\
"Halloween". Mayne also wrote a version of the \\
ballad "Helen of Kirkconnel". His verses were \\
admired by Walter Scott. Life. He was born at Dumfries \\
on 26 March 1759. Educated at the local grammar \\
school, he became a printer in the office of the \\
"Dumfries Journal". In 1782 he went with his family \\
to Glasgow, where he worked for five years in the \\
publishing house of the brothers Foulis. In 1787 \\
he settled in London, first as a printer, and then \\
as proprietor and joint editor of "The Star", an \\
evening paper, in which he placed his poems. He \\
died at Lisson Grove, London, 14 March 1836. Works. \\
Mayne wrote poetry in Dumfries, and after 1777 he \\
contributed poems to "Ruddiman\'s Weekly Magazine", \\
Edinburgh. Between 1807 and 1817 several of his \\
lyrics appeared in the "Gentleman\'s Magazine \\
\textcolor{brown}{\textbf{Question}} : What is John Mayne\'s occupation?\\
\textcolor{brown}{\textbf{Answer}} : Journalist \\
\midrule
\textbf{Model Response} \\
\textcolor{brown}{\textbf{Supported}} \\
} \\
\bottomrule[1.5pt]

\end{tabular}} \caption{Inference prompt for the support label extraction.} \label{tab:sup_prompt}
\end{table}

%% file: Tables/tb_data_sample.tex
\definecolor{metacolorscore}{HTML}{FAF7F0}

\begin{table}[h]
\centering
\resizebox{0.97\linewidth}{!}{
\begin{tabular}{l}

\toprule[1.5pt]
\rowcolor{metacolorscore} \makecell[l]{
\textbf{Query} \\
What is Anthony Sharps occupation? \\
\midrule
\textbf{Answer} \\
\text{[actor, actress, actors, actresses]} \\
\midrule
\textbf{Context} \\
1. British actor (1915–1984)  \\
Dennis Anthony John Sharp (16 June 1915 – 23 July 1984) \\
was an English actor, writer and director. Stage career. \\
Anthony Sharp was a graduate of the London Academy of \\
Music and Dramatic Art (LAMDA) and made his stage... \\
2. There he played Benedick in "Much Ado About Nothing" \\
in 1958 and Malvolio in "Twelfth Night" the following \\
year. Rejoining the company in the 1970s, he appeared \\
in such plays as "Love's Labour's Lost" and "The Man of...  \\
3. His credits included "Any Other Business" \\
(Westminster Theatre 1958),  "Caught Napping" (Piccadilly \\
Theatre 1959), "Wolf's Clothing" (Strand Theatre 1959), \\
"Billy Bunter Flies East" (Victoria Palace 1959), "The...  \\
4. His only starring role in a feature film was the \\
homicidal priest Father Xavier Meldrum in Pete Walker's \\
1975 horror picture "House of Mortal Sin".  His final \\
feature film, in which he played foreign secretary Lord Ambrose,...  \\
5. In 1974, he appeared as the vicar in the radio \\
version of "Steptoe and Son", and in 1978 he was both \\
Garkbit, the waiter in the Restaurant at the End of the \\
Universe , and The Great Prophet Zarquon in Fit the Fifth of the... \\
} \\
\bottomrule[1.5pt]
\end{tabular}}
\caption{Data sample with 5 relevant chunks. These examples include queries with general questions such that the answer can be inferred throughout the entire Wikipedia article.}
\label{tab:data_sample}
\end{table}

%% file: Tables/tb_rag_all.tex
\definecolor{metacolorab}{HTML}{F5F5F7}


\begin{table*}[t]
\centering
\resizebox{0.8\textwidth}{!}{
\begin{tabular}{c|c|cccc}
\toprule[1.5pt]

\makecell[c]{\textbf{LLM}} & \makecell[c]{\textbf{Retriever}} & \makecell[c]{\textbf{Noise} \\ \textbf{Vulnerability}($\downarrow$)} &
\makecell[c]{\textbf{Context} \\ \textbf{Acceptability} ($\uparrow$)} &
\makecell[c]{\textbf{Context} \\ \textbf{Insensitivity} ($\downarrow$)} &
\makecell[c]{\textbf{Context} \\ \textbf{Misintepretation} ($\downarrow$)} \\ \midrule[1.5pt]

\rowcolor{metacolorab} \multicolumn{6}{c}{\textit{\textbf{Top-1}}} \\
\midrule

\multirow{5}{*}{\makecell[l]{\textbf{GPT3.5}}} & \textbf{BGE-S} & 25.69 & 65.33 & 8.44 & 0.54 \\
 & \textbf{BGE-B} & 24.36 & 66.67 & 8.44 & 0.54 \\
 & \textbf{BGE-L} & 21.67 & 69.35 & 8.44 & 0.54 \\
 & \textbf{Contriever} & 42.26 & 48.76 & 8.44 & 0.54 \\
 & \textbf{NV} & \textbf{17.66} & \textbf{73.36} & 8.44 & 0.54 \\
 \midrule
 
\multirow{5}{*}{\makecell[l]{\textbf{GPT4o}}} & \textbf{BGE-S} & 22.05 & 69.09 & 8.29 & 0.57 \\
 & \textbf{BGE-B} & 20.57 & 70.57 & 8.29 & 0.57 \\
 & \textbf{BGE-L} & 18.83 & 72.31 & 8.29 & 0.57 \\
 & \textbf{Contriever} & 35.73 & 55.42 & 8.29 & 0.57 \\
 & \textbf{NV} & \textbf{15.38} & \textbf{75.76} & 8.29 & 0.57 \\
 \midrule
 
\multirow{5}{*}{\makecell[l]{\textbf{LLAMA2-7B}}} & \textbf{BGE-S} & 26.83 & 56.13 & 16.37 & 0.67 \\
 & \textbf{BGE-B} & 25.77 & 57.19 & 16.37 & 0.67 \\
 & \textbf{BGE-L} & 22.80 & 60.16 & 16.37 & 0.67 \\
 & \textbf{Contriever} & 47.63 & 35.33 & 16.37 & 0.67 \\
 & \textbf{NV} & \textbf{18.21} & \textbf{64.75} & 16.37 & 0.67 \\
 \midrule
 
\multirow{5}{*}{\makecell[l]{\textbf{Qwen2-7B}}} & \textbf{BGE-S} & 29.1 & 61.11 & 9.41 & 0.37 \\
 & \textbf{BGE-B} & 27.75 & 62.46 & 9.41 & 0.37 \\
 & \textbf{BGE-L} & 24.29 & 65.93 & 9.41 & 0.37 \\
 & \textbf{Contriever} & 51.14 & 39.07 & 9.41 & 0.37 \\
 & \textbf{NV} & \textbf{19.04} & \textbf{71.17} & 9.41 & 0.37 \\

\bottomrule[1.5pt]
\end{tabular}
}
\caption{Top1 performance for RAG systems}
\label{tab:rag_all_top1}
\end{table*}


\begin{table*}[t]
\centering
\resizebox{0.8\textwidth}{!}{
\begin{tabular}{c|c|cccc}
\toprule[1.5pt]

\makecell[c]{\textbf{LLM}} & \makecell[c]{\textbf{Retriever}} & \makecell[c]{\textbf{Noise} \\ \textbf{Vulnerability}($\downarrow$)} &
\makecell[c]{\textbf{Context} \\ \textbf{Acceptability} ($\uparrow$)} &
\makecell[c]{\textbf{Context} \\ \textbf{Insensitivity} ($\downarrow$)} &
\makecell[c]{\textbf{Context} \\ \textbf{Misintepretation} ($\downarrow$)} \\ \midrule[1.5pt]

\rowcolor{metacolorab} \multicolumn{6}{c}{\textit{\textbf{Top-3}}} \\
\midrule[1.5pt]

\multirow{5}{*}{\makecell[l]{\textbf{GPT3.5}}} & \textbf{BGE-S} & 18.52 & 72.50 & 8.44 & 0.54 \\
 & \textbf{BGE-B} & 17.39 & 73.63 & 8.44 & 0.54 \\
 & \textbf{BGE-L} & 16.00 & 75.02 & 8.44 & 0.54 \\
 & \textbf{Contriever} & 30.91 & 60.11 & 8.44 & 0.54 \\
 & \textbf{NV} & \textbf{13.16} & \textbf{77.87} & 8.44 & 0.54 \\
 \midrule
\multirow{5}{*}{\makecell[l]{\textbf{GPT4o}}} & \textbf{BGE-S} & 14.02 & 77.12 & 8.29 & 0.57 \\
 & \textbf{BGE-B} & 13.67 & 77.47 & 8.29 & 0.57 \\
 & \textbf{BGE-L} & 12.38 & 78.77 & 8.29 & 0.57 \\
 & \textbf{Contriever} & 25.94 & 65.20 & 8.29 & 0.57 \\
 & \textbf{NV} & \textbf{10.75} & \textbf{80.39} & 8.29 & 0.57 \\
 \midrule
\multirow{5}{*}{\makecell[l]{\textbf{LLAMA2-7B}}} & \textbf{BGE-S} & 22.32 & 60.64 & 16.37 & 0.67 \\
 & \textbf{BGE-B} & 21.72 & 61.23 & 16.37 & 0.67 \\
 & \textbf{BGE-L} & 20.32 & 62.63 & 16.37 & 0.67 \\
 & \textbf{Contriever} & 36.56 & 46.40 & 16.37 & 0.67 \\
 & \textbf{NV} & \textbf{16.81} & \textbf{66.15} & 16.37 & 0.67 \\
 \midrule
\multirow{5}{*}{\makecell[l]{\textbf{Qwen2-7B}}} & \textbf{BGE-S} & 21.88 & 68.33 & 9.41 & 0.37 \\
 & \textbf{BGE-B} & 21.04 & 69.18 & 9.41 & 0.37 \\
 & \textbf{BGE-L} & 19.33 & 70.88 & 9.41 & 0.37 \\
 & \textbf{Contriever} & 37.09 & 53.13 & 9.41 & 0.37 \\
 & \textbf{NV} & \textbf{15.83} & \textbf{74.39} & 9.41 & 0.37 \\

\bottomrule[1.5pt]
\end{tabular}
}
\caption{Top3 performance for RAG systems}
\label{tab:rag_all_top3}
\end{table*}


\begin{table*}[t]
\centering
\resizebox{0.8\textwidth}{!}{
\begin{tabular}{c|c|cccc}
\toprule[1.5pt]

\makecell[c]{\textbf{LLM}} & \makecell[c]{\textbf{Retriever}} & \makecell[c]{\textbf{Noise} \\ \textbf{Vulnerability}($\downarrow$)} &
\makecell[c]{\textbf{Context} \\ \textbf{Acceptability} ($\uparrow$)} &
\makecell[c]{\textbf{Context} \\ \textbf{Insensitivity} ($\downarrow$)} &
\makecell[c]{\textbf{Contect} \\ \textbf{Misintepretation} ($\downarrow$)} \\ \midrule[1.5pt]

\rowcolor{metacolorab} \multicolumn{6}{c}{\textit{\textbf{Top-5}}} \\
\midrule[1.5pt]

\multirow{5}{*}{\makecell[l]{\textbf{GPT3.5}}} & \textbf{BGE-S} & 17.36 & 73.66 & 8.44 & 0.54 \\
 & \textbf{BGE-B} & 16.42 & 74.60 & 8.44 & 0.54 \\
 & \textbf{BGE-L} & 15.36 & 75.66 & 8.44 & 0.54 \\
 & \textbf{Contriever} & 27.45 & 63.57 & 8.44 & 0.54 \\
 & \textbf{NV} & \textbf{13.21} & \textbf{77.81} & 8.44 & 0.54 \\
 \midrule
\multirow{5}{*}{\makecell[l]{\textbf{GPT4o}}} & \textbf{BGE-S} & 13.17 & 77.97 & 8.29 & 0.57 \\
 & \textbf{BGE-B} & 12.79 & 78.35 & 8.29 & 0.57 \\
 & \textbf{BGE-L} & 11.78 & 79.36 & 8.29 & 0.57 \\
 & \textbf{Contriever} & 22.65 & 68.49 & 8.29 & 0.57 \\
 & \textbf{NV} & \textbf{10.64} & \textbf{80.50} & 8.29 & 0.57 \\
 \midrule
\multirow{5}{*}{\makecell[l]{\textbf{LLAMA2-7B}}} & \textbf{BGE-S} & 24.12 & 58.84 & 16.37 & 0.67 \\
 & \textbf{BGE-B} & 24.08 & 58.88 & 16.37 & 0.67 \\
 & \textbf{BGE-L} & 23.19 & 59.77 & 16.37 & 0.67 \\
 & \textbf{Contriever} & 36.78 & 46.17 & 16.37 & 0.67 \\
 & \textbf{NV} & \textbf{19.93} & \textbf{63.03} & 16.37 & 0.67 \\
 \midrule
\multirow{5}{*}{\makecell[l]{\textbf{Qwen2-7B}}} & \textbf{BGE-S} & 22.5 & 67.71 & 9.41 & 0.37 \\
 & \textbf{BGE-B} & 21.82 & 68.40 & 9.41 & 0.37 \\
 & \textbf{BGE-L} & 20.15 & 70.06 & 9.41 & 0.37 \\
 & \textbf{Contriever} & 34.46 & 55.76 & 9.41 & 0.37 \\
 & \textbf{NV} & \textbf{17.61} & \textbf{72.60} & 9.41 & 0.37 \\

\bottomrule[1.5pt]
\end{tabular}
}
\caption{Top5 performance for RAG systems}
\label{tab:rag_all_top5}
\end{table*}

%% file: MIRAGE.bbl
\begin{thebibliography}{42}
\providecommand{\natexlab}[1]{#1}

\bibitem[{Achiam et~al.(2023)Achiam, Adler, Agarwal, Ahmad, Akkaya, Aleman, Almeida, Altenschmidt, Altman, Anadkat et~al.}]{achiam2023gpt}
Josh Achiam, Steven Adler, Sandhini Agarwal, Lama Ahmad, Ilge Akkaya, Florencia~Leoni Aleman, Diogo Almeida, Janko Altenschmidt, Sam Altman, Shyamal Anadkat, et~al. 2023.
\newblock Gpt-4 technical report.
\newblock \emph{arXiv preprint arXiv:2303.08774}.

\bibitem[{Bajaj et~al.(2016)Bajaj, Campos, Craswell, Deng, Gao, Liu, Majumder, McNamara, Mitra, Nguyen et~al.}]{bajaj2016ms}
Payal Bajaj, Daniel Campos, Nick Craswell, Li~Deng, Jianfeng Gao, Xiaodong Liu, Rangan Majumder, Andrew McNamara, Bhaskar Mitra, Tri Nguyen, et~al. 2016.
\newblock Ms marco: A human generated machine reading comprehension dataset.
\newblock \emph{arXiv preprint arXiv:1611.09268}.

\bibitem[{Chen et~al.(2024{\natexlab{a}})Chen, Xiao, Zhang, Luo, Lian, and Liu}]{chen2024bge}
Jianlv Chen, Shitao Xiao, Peitian Zhang, Kun Luo, Defu Lian, and Zheng Liu. 2024{\natexlab{a}}.
\newblock Bge m3-embedding: Multi-lingual, multi-functionality, multi-granularity text embeddings through self-knowledge distillation.
\newblock \emph{arXiv preprint arXiv:2402.03216}.

\bibitem[{Chen et~al.(2024{\natexlab{b}})Chen, Lin, Han, and Sun}]{chen2024benchmarking}
Jiawei Chen, Hongyu Lin, Xianpei Han, and Le~Sun. 2024{\natexlab{b}}.
\newblock Benchmarking large language models in retrieval-augmented generation.
\newblock In \emph{Proceedings of the AAAI Conference on Artificial Intelligence}, volume~38, pages 17754--17762.

\bibitem[{Cohere(2024)}]{cohere2024command}
Cohere. 2024.
\newblock Command r models.
\newblock \url{https://cohere.com/command}.
\newblock Accessed: 2024-03-01.

\bibitem[{Dua et~al.(2019)Dua, Wang, Dasigi, Stanovsky, Singh, and Gardner}]{dua2019drop}
Dheeru Dua, Yizhong Wang, Pradeep Dasigi, Gabriel Stanovsky, Sameer Singh, and Matt Gardner. 2019.
\newblock Drop: A reading comprehension benchmark requiring discrete reasoning over paragraphs.
\newblock \emph{arXiv preprint arXiv:1903.00161}.

\bibitem[{Dubey et~al.(2024)Dubey, Jauhri, Pandey, Kadian, Al-Dahle, Letman, Mathur, Schelten, Yang, Fan et~al.}]{dubey2024llama}
Abhimanyu Dubey, Abhinav Jauhri, Abhinav Pandey, Abhishek Kadian, Ahmad Al-Dahle, Aiesha Letman, Akhil Mathur, Alan Schelten, Amy Yang, Angela Fan, et~al. 2024.
\newblock The llama 3 herd of models.
\newblock \emph{arXiv preprint arXiv:2407.21783}.

\bibitem[{Es et~al.(2023)Es, James, Espinosa-Anke, and Schockaert}]{es2023ragas}
Shahul Es, Jithin James, Luis Espinosa-Anke, and Steven Schockaert. 2023.
\newblock Ragas: Automated evaluation of retrieval augmented generation.
\newblock \emph{arXiv preprint arXiv:2309.15217}.

\bibitem[{Fabbri et~al.(2021)Fabbri, Kry{\'s}ci{\'n}ski, McCann, Xiong, Socher, and Radev}]{fabbri2021summeval}
Alexander~R Fabbri, Wojciech Kry{\'s}ci{\'n}ski, Bryan McCann, Caiming Xiong, Richard Socher, and Dragomir Radev. 2021.
\newblock Summeval: Re-evaluating summarization evaluation.
\newblock \emph{Transactions of the Association for Computational Linguistics}, 9:391--409.

\bibitem[{Fan et~al.(2024)Fan, Ding, Ning, Wang, Li, Yin, Chua, and Li}]{fan2024survey}
Wenqi Fan, Yujuan Ding, Liangbo Ning, Shijie Wang, Hengyun Li, Dawei Yin, Tat-Seng Chua, and Qing Li. 2024.
\newblock A survey on rag meeting llms: Towards retrieval-augmented large language models.
\newblock In \emph{Proceedings of the 30th ACM SIGKDD Conference on Knowledge Discovery and Data Mining}, pages 6491--6501.

\bibitem[{Gao et~al.(2023)Gao, Xiong, Gao, Jia, Pan, Bi, Dai, Sun, and Wang}]{gao2023retrieval}
Yunfan Gao, Yun Xiong, Xinyu Gao, Kangxiang Jia, Jinliu Pan, Yuxi Bi, Yi~Dai, Jiawei Sun, and Haofen Wang. 2023.
\newblock Retrieval-augmented generation for large language models: A survey.
\newblock \emph{arXiv preprint arXiv:2312.10997}.

\bibitem[{Hofst{\"a}tter et~al.(2023)Hofst{\"a}tter, Chen, Raman, and Zamani}]{hofstatter2023fid}
Sebastian Hofst{\"a}tter, Jiecao Chen, Karthik Raman, and Hamed Zamani. 2023.
\newblock Fid-light: Efficient and effective retrieval-augmented text generation.
\newblock In \emph{Proceedings of the 46th International ACM SIGIR Conference on Research and Development in Information Retrieval}, pages 1437--1447.

\bibitem[{Hsieh et~al.(2023)Hsieh, Chen, Li, Fujii, Ratner, Lee, Krishna, and Pfister}]{hsieh2023tool}
Cheng-Yu Hsieh, Si-An Chen, Chun-Liang Li, Yasuhisa Fujii, Alexander Ratner, Chen-Yu Lee, Ranjay Krishna, and Tomas Pfister. 2023.
\newblock Tool documentation enables zero-shot tool-usage with large language models.
\newblock \emph{arXiv preprint arXiv:2308.00675}.

\bibitem[{Izacard et~al.(2021)Izacard, Caron, Hosseini, Riedel, Bojanowski, Joulin, and Grave}]{izacard2021unsupervised}
Gautier Izacard, Mathilde Caron, Lucas Hosseini, Sebastian Riedel, Piotr Bojanowski, Armand Joulin, and Edouard Grave. 2021.
\newblock Unsupervised dense information retrieval with contrastive learning.
\newblock \emph{arXiv preprint arXiv:2112.09118}.

\bibitem[{Izacard and Grave(2021)}]{izacard-grave-2021-leveraging}
Gautier Izacard and Edouard Grave. 2021.
\newblock \href {https://doi.org/10.18653/v1/2021.eacl-main.74} {Leveraging passage retrieval with generative models for open domain question answering}.
\newblock In \emph{Proceedings of the 16th Conference of the European Chapter of the Association for Computational Linguistics: Main Volume}, pages 874--880, Online. Association for Computational Linguistics.

\bibitem[{Ji et~al.(2023)Ji, Yu, Xu, Lee, Ishii, and Fung}]{ji2023towards}
Ziwei Ji, Tiezheng Yu, Yan Xu, Nayeon Lee, Etsuko Ishii, and Pascale Fung. 2023.
\newblock Towards mitigating llm hallucination via self reflection.
\newblock In \emph{Findings of the Association for Computational Linguistics: EMNLP 2023}, pages 1827--1843.

\bibitem[{Joshi et~al.(2017)Joshi, Choi, Weld, and Zettlemoyer}]{joshi2017triviaqa}
Mandar Joshi, Eunsol Choi, Daniel~S Weld, and Luke Zettlemoyer. 2017.
\newblock Triviaqa: A large scale distantly supervised challenge dataset for reading comprehension.
\newblock \emph{arXiv preprint arXiv:1705.03551}.

\bibitem[{Karpukhin et~al.(2020)Karpukhin, O{\u{g}}uz, Min, Lewis, Wu, Edunov, Chen, and Yih}]{karpukhin2020dense}
Vladimir Karpukhin, Barlas O{\u{g}}uz, Sewon Min, Patrick Lewis, Ledell Wu, Sergey Edunov, Danqi Chen, and Wen-tau Yih. 2020.
\newblock Dense passage retrieval for open-domain question answering.
\newblock \emph{arXiv preprint arXiv:2004.04906}.

\bibitem[{Kasai et~al.(2023)Kasai, Sakaguchi, takahashi, Le~Bras, Asai, Yu, Radev, Smith, Choi, and Inui}]{realtimeqa}
Jungo Kasai, Keisuke Sakaguchi, yoichi takahashi, Ronan Le~Bras, Akari Asai, Xinyan Yu, Dragomir Radev, Noah~A Smith, Yejin Choi, and Kentaro Inui. 2023.
\newblock \href {https://proceedings.neurips.cc/paper_files/paper/2023/file/9941624ef7f867a502732b5154d30cb7-Paper-Datasets_and_Benchmarks.pdf} {Realtime qa: What\textquotesingle s the answer right now?}
\newblock In \emph{Advances in Neural Information Processing Systems}, volume~36, pages 49025--49043. Curran Associates, Inc.

\bibitem[{Kwiatkowski et~al.(2019)Kwiatkowski, Palomaki, Redfield, Collins, Parikh, Alberti, Epstein, Polosukhin, Devlin, Lee et~al.}]{kwiatkowski2019natural}
Tom Kwiatkowski, Jennimaria Palomaki, Olivia Redfield, Michael Collins, Ankur Parikh, Chris Alberti, Danielle Epstein, Illia Polosukhin, Jacob Devlin, Kenton Lee, et~al. 2019.
\newblock Natural questions: a benchmark for question answering research.
\newblock \emph{Transactions of the Association for Computational Linguistics}, 7:453--466.

\bibitem[{Kwon et~al.(2023)Kwon, Li, Zhuang, Sheng, Zheng, Yu, Gonzalez, Zhang, and Stoica}]{kwon2023efficient}
Woosuk Kwon, Zhuohan Li, Siyuan Zhuang, Ying Sheng, Lianmin Zheng, Cody~Hao Yu, Joseph Gonzalez, Hao Zhang, and Ion Stoica. 2023.
\newblock Efficient memory management for large language model serving with pagedattention.
\newblock In \emph{Proceedings of the 29th Symposium on Operating Systems Principles}, pages 611--626.

\bibitem[{Lee et~al.(2024)Lee, Roy, Xu, Raiman, Shoeybi, Catanzaro, and Ping}]{lee2024nv}
Chankyu Lee, Rajarshi Roy, Mengyao Xu, Jonathan Raiman, Mohammad Shoeybi, Bryan Catanzaro, and Wei Ping. 2024.
\newblock Nv-embed: Improved techniques for training llms as generalist embedding models.
\newblock \emph{arXiv preprint arXiv:2405.17428}.

\bibitem[{Lewis et~al.(2020)Lewis, Perez, Piktus, Petroni, Karpukhin, Goyal, K{\"u}ttler, Lewis, Yih, Rockt{\"a}schel et~al.}]{lewis2020retrieval}
Patrick Lewis, Ethan Perez, Aleksandra Piktus, Fabio Petroni, Vladimir Karpukhin, Naman Goyal, Heinrich K{\"u}ttler, Mike Lewis, Wen-tau Yih, Tim Rockt{\"a}schel, et~al. 2020.
\newblock Retrieval-augmented generation for knowledge-intensive nlp tasks.
\newblock \emph{Advances in Neural Information Processing Systems}, 33:9459--9474.

\bibitem[{Li et~al.(2021)Li, Li, Shang, Jiang, Liu, Sun, Ji, and Liu}]{li2021hopretriever}
Shaobo Li, Xiaoguang Li, Lifeng Shang, Xin Jiang, Qun Liu, Chengjie Sun, Zhenzhou Ji, and Bingquan Liu. 2021.
\newblock Hopretriever: Retrieve hops over wikipedia to answer complex questions.
\newblock In \emph{Proceedings of the AAAI conference on artificial intelligence}, volume~35, pages 13279--13287.

\bibitem[{Liu et~al.(2023)Liu, Huang, Li, Chen, Zhou, Meng, Zhou, and Sun}]{liu2023recall}
Yi~Liu, Lianzhe Huang, Shicheng Li, Sishuo Chen, Hao Zhou, Fandong Meng, Jie Zhou, and Xu~Sun. 2023.
\newblock Recall: A benchmark for llms robustness against external counterfactual knowledge.
\newblock \emph{arXiv preprint arXiv:2311.08147}.

\bibitem[{Lu et~al.(2023)Lu, Peng, Cheng, Galley, Chang, Wu, Zhu, and Gao}]{Chameleon}
Pan Lu, Baolin Peng, Hao Cheng, Michel Galley, Kai-Wei Chang, Ying~Nian Wu, Song-Chun Zhu, and Jianfeng Gao. 2023.
\newblock \href {https://proceedings.neurips.cc/paper_files/paper/2023/file/871ed095b734818cfba48db6aeb25a62-Paper-Conference.pdf} {Chameleon: Plug-and-play compositional reasoning with large language models}.
\newblock In \emph{Advances in Neural Information Processing Systems}, volume~36, pages 43447--43478. Curran Associates, Inc.

\bibitem[{Mallen et~al.(2022)Mallen, Asai, Zhong, Das, Khashabi, and Hajishirzi}]{mallen2022not}
Alex Mallen, Akari Asai, Victor Zhong, Rajarshi Das, Daniel Khashabi, and Hannaneh Hajishirzi. 2022.
\newblock When not to trust language models: Investigating effectiveness of parametric and non-parametric memories.
\newblock \emph{arXiv preprint arXiv:2212.10511}.

\bibitem[{Mallen et~al.(2023)Mallen, Asai, Zhong, Das, Khashabi, and Hajishirzi}]{mallen-etal-2023-trust}
Alex Mallen, Akari Asai, Victor Zhong, Rajarshi Das, Daniel Khashabi, and Hannaneh Hajishirzi. 2023.
\newblock \href {https://doi.org/10.18653/v1/2023.acl-long.546} {When not to trust language models: Investigating effectiveness of parametric and non-parametric memories}.
\newblock In \emph{Proceedings of the 61st Annual Meeting of the Association for Computational Linguistics (Volume 1: Long Papers)}, pages 9802--9822, Toronto, Canada. Association for Computational Linguistics.

\bibitem[{Neelakantan et~al.(2022)Neelakantan, Xu, Puri, Radford, Han, Tworek, Yuan, Tezak, Kim, Hallacy et~al.}]{neelakantan2022text}
Arvind Neelakantan, Tao Xu, Raul Puri, Alec Radford, Jesse~Michael Han, Jerry Tworek, Qiming Yuan, Nikolas Tezak, Jong~Wook Kim, Chris Hallacy, et~al. 2022.
\newblock Text and code embeddings by contrastive pre-training.
\newblock \emph{arXiv preprint arXiv:2201.10005}.

\bibitem[{Ru et~al.(2024)Ru, Qiu, Hu, Zhang, Shi, Chang, Cheng, Wang, Sun, Li et~al.}]{ru2024ragchecker}
Dongyu Ru, Lin Qiu, Xiangkun Hu, Tianhang Zhang, Peng Shi, Shuaichen Chang, Jiayang Cheng, Cunxiang Wang, Shichao Sun, Huanyu Li, et~al. 2024.
\newblock Ragchecker: A fine-grained framework for diagnosing retrieval-augmented generation.
\newblock \emph{arXiv preprint arXiv:2408.08067}.

\bibitem[{Saad-Falcon et~al.(2023)Saad-Falcon, Khattab, Potts, and Zaharia}]{saad2023ares}
Jon Saad-Falcon, Omar Khattab, Christopher Potts, and Matei Zaharia. 2023.
\newblock Ares: An automated evaluation framework for retrieval-augmented generation systems.
\newblock \emph{arXiv preprint arXiv:2311.09476}.

\bibitem[{Touvron et~al.(2023)Touvron, Martin, Stone, Albert, Almahairi, Babaei, Bashlykov, Batra, Bhargava, Bhosale et~al.}]{touvron2023llama}
Hugo Touvron, Louis Martin, Kevin Stone, Peter Albert, Amjad Almahairi, Yasmine Babaei, Nikolay Bashlykov, Soumya Batra, Prajjwal Bhargava, Shruti Bhosale, et~al. 2023.
\newblock Llama 2: Open foundation and fine-tuned chat models.
\newblock \emph{arXiv preprint arXiv:2307.09288}.

\bibitem[{Vu et~al.(2023)Vu, Iyyer, Wang, Constant, Wei, Wei, Tar, Sung, Zhou, Le et~al.}]{vu2023freshllms}
Tu~Vu, Mohit Iyyer, Xuezhi Wang, Noah Constant, Jerry Wei, Jason Wei, Chris Tar, Yun-Hsuan Sung, Denny Zhou, Quoc Le, et~al. 2023.
\newblock Freshllms: Refreshing large language models with search engine augmentation.
\newblock \emph{arXiv preprint arXiv:2310.03214}.

\bibitem[{Wang et~al.(2024)Wang, Yang, Huang, Yang, Majumder, and Wei}]{wang2024multilingual}
Liang Wang, Nan Yang, Xiaolong Huang, Linjun Yang, Rangan Majumder, and Furu Wei. 2024.
\newblock Multilingual e5 text embeddings: A technical report.
\newblock \emph{arXiv preprint arXiv:2402.05672}.

\bibitem[{Xie et~al.(2024)Xie, Zhang, Chen, Lou, and Su}]{xie2024adaptive}
Jian Xie, Kai Zhang, Jiangjie Chen, Renze Lou, and Yu~Su. 2024.
\newblock \href {https://openreview.net/forum?id=auKAUJZMO6} {Adaptive chameleon or stubborn sloth: Revealing the behavior of large language models in knowledge conflicts}.
\newblock In \emph{The Twelfth International Conference on Learning Representations}.

\bibitem[{Yang et~al.(2024)Yang, Yang, Hui, Zheng, Yu, Zhou, Li, Li, Liu, Huang et~al.}]{yang2024qwen2}
An~Yang, Baosong Yang, Binyuan Hui, Bo~Zheng, Bowen Yu, Chang Zhou, Chengpeng Li, Chengyuan Li, Dayiheng Liu, Fei Huang, et~al. 2024.
\newblock Qwen2 technical report.
\newblock \emph{arXiv preprint arXiv:2407.10671}.

\bibitem[{Yang et~al.(2015)Yang, Yih, and Meek}]{yang2015wikiqa}
Yi~Yang, Wen-tau Yih, and Christopher Meek. 2015.
\newblock Wikiqa: A challenge dataset for open-domain question answering.
\newblock In \emph{Proceedings of the 2015 conference on empirical methods in natural language processing}, pages 2013--2018.

\bibitem[{Yang et~al.(2018)Yang, Qi, Zhang, Bengio, Cohen, Salakhutdinov, and Manning}]{yang2018hotpotqa}
Zhilin Yang, Peng Qi, Saizheng Zhang, Yoshua Bengio, William~W Cohen, Ruslan Salakhutdinov, and Christopher~D Manning. 2018.
\newblock Hotpotqa: A dataset for diverse, explainable multi-hop question answering.
\newblock \emph{arXiv preprint arXiv:1809.09600}.

\bibitem[{Yu et~al.(2023{\natexlab{a}})Yu, Iter, Wang, Xu, Ju, Sanyal, Zhu, Zeng, and Jiang}]{yu2023generate}
Wenhao Yu, Dan Iter, Shuohang Wang, Yichong Xu, Mingxuan Ju, Soumya Sanyal, Chenguang Zhu, Michael Zeng, and Meng Jiang. 2023{\natexlab{a}}.
\newblock \href {https://openreview.net/forum?id=fB0hRu9GZUS} {Generate rather than retrieve: Large language models are strong context generators}.
\newblock In \emph{The Eleventh International Conference on Learning Representations}.

\bibitem[{Yu et~al.(2023{\natexlab{b}})Yu, Jiang, Clark, and Sabharwal}]{yu2023ifqa}
Wenhao Yu, Meng Jiang, Peter Clark, and Ashish Sabharwal. 2023{\natexlab{b}}.
\newblock Ifqa: A dataset for open-domain question answering under counterfactual presuppositions.
\newblock \emph{arXiv preprint arXiv:2305.14010}.

\bibitem[{Zhang et~al.(2024)Zhang, Zhang, Long, Xie, Dai, Tang, Lin, Yang, Xie, Huang et~al.}]{zhang2024mgte}
Xin Zhang, Yanzhao Zhang, Dingkun Long, Wen Xie, Ziqi Dai, Jialong Tang, Huan Lin, Baosong Yang, Pengjun Xie, Fei Huang, et~al. 2024.
\newblock mgte: Generalized long-context text representation and reranking models for multilingual text retrieval.
\newblock \emph{arXiv preprint arXiv:2407.19669}.

\bibitem[{Zhao et~al.(2024)Zhao, Zhang, Yu, Wang, Geng, Fu, Yang, Zhang, and Cui}]{corr_survey}
Penghao Zhao, Hailin Zhang, Qinhan Yu, Zhengren Wang, Yunteng Geng, Fangcheng Fu, Ling Yang, Wentao Zhang, and Bin Cui. 2024.
\newblock \href {https://doi.org/10.48550/arXiv.2402.19473} {Retrieval-augmented generation for ai-generated content: A survey}.
\newblock \emph{CoRR}, abs/2402.19473.

\end{thebibliography}
